\documentclass[journal,twocolumn,10pt]{IEEEtran}
\usepackage{graphicx,times,amsmath,amssymb,cite,subfigure,url}
\usepackage[lined,ruled,commentsnumbered]{algorithm2e}
\IEEEoverridecommandlockouts \allowdisplaybreaks 

\begin{document}

\title{Driver Drowsiness Estimation from EEG Signals Using Online Weighted Adaptation Regularization for Regression (OwARR)}

\author{\IEEEauthorblockN{Dongrui Wu\IEEEauthorrefmark{1}, \textit{Senior Member, IEEE}, Vernon J. Lawhern\IEEEauthorrefmark{2}\IEEEauthorrefmark{3}, \textit{Member, IEEE},\\ Stephen Gordon\IEEEauthorrefmark{4}, Brent J. Lance\IEEEauthorrefmark{2}, \textit{Senior Member, IEEE}, Chin-Teng Lin\IEEEauthorrefmark{5}\IEEEauthorrefmark{6}, \textit{Fellow, IEEE}} \\
\IEEEauthorblockA{\IEEEauthorrefmark{1}DataNova, NY USA}\\
\IEEEauthorblockA{\IEEEauthorrefmark{2}Human Research and Engineering Directorate, U.S. Army Research Laboratory, Aberdeen Proving Ground, MD USA}\\
\IEEEauthorblockA{\IEEEauthorrefmark{3}Department of Computer Science, University of Texas at San Antonio, San Antonio, TX USA}\\
\IEEEauthorblockA{\IEEEauthorrefmark{4}DCS Corp, Alexandria, VA USA}\\
\IEEEauthorblockA{\IEEEauthorrefmark{5}Faculty of Engineering and Information Technology, University of Technology, Sydney, Australia}\\
\IEEEauthorblockA{\IEEEauthorrefmark{6}Brain Research Center, National Chiao-Tung University, Hsinchu, Taiwan}\\
E-mail: drwu09@gmail.com, vernon.j.lawhern.civ@mail.mil, sgordon@dcscorp.com, brent.j.lance.civ@mail.mil, Chin-Teng.Lin@uts.edu.au}
\maketitle

\begin{abstract}
One big challenge that hinders the transition of brain-computer interfaces (BCIs) from laboratory settings to real-life applications is the availability of high-performance and robust learning algorithms that can effectively handle individual differences, i.e., algorithms that can be applied to a new subject with zero or very little subject-specific calibration data. Transfer learning and domain adaptation have been extensively used for this purpose. However, most previous works focused on classification problems. This paper considers an important regression problem in BCI, namely, online driver drowsiness estimation from EEG signals. By integrating fuzzy sets with domain adaptation, we propose a novel online weighted adaptation regularization for regression (OwARR) algorithm to reduce the amount of subject-specific calibration data, and also a source domain selection (SDS) approach to save about half of the computational cost of OwARR. Using a simulated driving dataset with 15 subjects, we show that OwARR and OwARR-SDS can achieve significantly smaller estimation errors than several other approaches. We also provide comprehensive analyses on the robustness of OwARR and OwARR-SDS.
\end{abstract}


\section{Introduction}

Brain computer interfaces (BCIs) \cite{Muhl2014,Erp2012,Lance2012,Makeig2012,Wolpaw2012} have attracted rapidly increasing research interest in the last decade, thanks to recent advances in neurosciences, wearable/mobile biosensors, and analytics. However, there are still many challenges in their transition from laboratory settings to real-life applications, including the reliability and convenience of the sensing hardware \cite{Liao2012}, and the availability of high-performance and robust algorithms for signal analysis and interpretation that can effectively handle individual differences and non-stationarity \cite{Lotte2015,Makeig2012,Wang2015,Jayaram2016}. This paper focuses on the last challenge, more specifically, how to generalize a BCI algorithm to a new subject, with zero or very little subject-specific calibration data.

Transfer learning (TL) \cite{Pan2010}, which improves learning in a new task by leveraging data or knowledge from other relevant tasks, represents a promising solution to the above challenge. Many TL approaches have been proposed for BCI applications \cite{Wang2015}, including: 1)
\textit{feature representation transfer} \cite{Kang2009,Devlaminck2011,Samek2013,Spuler2012}, which encodes the knowledge across different tasks as features; 2) \textit{instance transfer} \cite{Li2010,Li2009,drwuTL2011,drwuPLOS2013}, which uses certain parts of the data from other tasks to help the learning for the current task; and, 3) \textit{classifier transfer}, which includes domain adaptation (DA) \cite{Vidaurre2011,Bamdadian2013,Spuler2012}, ensemble learning \cite{Tu2012,Tu2012b}, and their combinations \cite{drwuSMC2015,drwuACII2015,drwuaBCI2015}.

However, most of the above TL approaches consider only BCI classification problems. Reducing the calibration data requirement in BCI regression problems has been largely under-studied. One example is online driver drowsiness estimation from EEG signals, which will be investigated in this paper. This is a very important problem because drowsy driving is among the most important causes of road crashes, following only to alcohol, speeding, and inattention \cite{Sagberg2004}. According to the National Highway Traffic Safety Administration \cite{NHTSA2011}, 2.5\% of fatal motor vehicle crashes (on average 886/year in the U.S.) and 2.5\% of fatalities (on average 1,004/year in the U.S.) between 2005 and 2009 involved drowsy driving. However, to our best knowledge, there have been only two works \cite{Wei2015,drwuaBCI2015} on TL for drowsiness estimation. Wei et al. \cite{Wei2015} showed that selective TL, which selectively turns TL on or off based the level of session generalizability, can achieve better estimation performance than approaches that always turn TL on or off. Wu et al. \cite{drwuaBCI2015} proposed a domain adaptation with model fusion (DAMF) approach for drowsiness estimation. By making use of data from other subjects in a DA framework, DAMF requires very little subject-specific calibration data, which significantly increases its real-world applicability.

In this paper, by making use of fuzzy sets (FSs) \cite{Zadeh1965}, we extend our earlier work on online weighted adaptation regularization \cite{drwuACII2015} from classification to regression to estimate driver drowsiness online from EEG signals. We show that our two proposed algorithms can achieve significantly better estimation performance than the DAMF and two other baseline approaches.

The remainder of the paper is organized as follows: Section~\ref{sect:OwARR} introduces the details of the proposed online weighted adaptation regularization for regression (OwARR) algorithm. Section~\ref{sect:OwARR-SDS} further introduces a source domain selection (SDS) approach to save the computational cost of OwARR. Section~\ref{sect:experiments} presents experimental results and performance comparisons of OwARR and OwARR-SDS with three other approaches. Finally, Section~\ref{sect:conclusions} draws conclusions and points out future research directions.

\section{Online Weighted Adaptation Regularization for Regression (OwARR)}\label{sect:OwARR}

In \cite{drwuACII2015} we have defined two types of calibration in BCI:
\begin{enumerate}
\item \emph{Offline calibration}, in which a pool of unlabeled EEG epochs have been obtained \emph{a priori}, and a subject or an oracle is queried to label some of these epochs, which are then used to train a model to label the remaining epochs in the pool.
\item \emph{Online calibration}, in which some labeled EEG epochs are obtained on-the-fly, and then a model is trained from them for future (unseen) EEG epochs.
\end{enumerate}
The major different between them is that, for offline calibration, the unlabeled EEG epochs can be used to help design the model (e.g., semi-supervised learning), whereas in online calibration there are no unlabeled EEG epochs. Additionally, in offline calibration we can query any epoch in the pool for the label, but in online calibration usually the sequence of the epochs is pre-determined and the subject or oracle has little control on which epochs to see next.

We only consider online calibration in this paper. This section introduces the OwARR algorithm, which extends the online weighted adaptation regularization algorithm \cite{drwuACII2015} from classification to regression, by making use of FSs.

\subsection{Problem Definition}

A domain \cite{Pan2010,Long2014} $\mathcal{D}$ in TL consists of a $d$-dimensional feature space $\mathcal{X}$ and a marginal probability distribution $P(\mathbf{x})$, i.e., $\mathcal{D}=\{\mathcal{X},P(\mathbf{x})\}$, where $\mathbf{x}\in \mathcal{X}$. Two domains $\mathcal{D}^z$ and $\mathcal{D}^t$ are different if $\mathcal{X}^z\neq \mathcal{X}^t$, and/or $P^z(\mathbf{x})\neq P^t(\mathbf{x})$.

A task \cite{Pan2010,Long2014} $\mathcal{T}$ in TL consists of an output space $\mathcal{Y}$ and a conditional probability distribution $Q(y|\mathbf{x})$. Two tasks $\mathcal{T}^z$ and $\mathcal{T}^t$ are different if $\mathcal{Y}^z\neq \mathcal{Y}^t$, or $Q^z(y|\mathbf{x})\neq Q^t(y|\mathbf{x})$.

Given the $z^{\mathrm{th}}$ source domain $\mathcal{D}^z$ with $n_z$ samples $(\mathbf{x}_i^z,y_i^z)$, $i=1,...,n_z$, and a target domain $\mathcal{D}^t$ with $m$ calibration samples $(\mathbf{x}_j^t,y_j^t)$, $j=1,...,m$, DA aims to learn a target prediction function $f(\mathbf{x}): \mathbf{x} \mapsto y$ with low expected error on $\mathcal{D}^t$, under the assumptions that $\mathcal{X}^z=\mathcal{X}^t$, $\mathcal{Y}^z=\mathcal{Y}^t$, $P^z(\mathbf{x})\neq P^t(\mathbf{x})$, and $Q^z(y|\mathbf{x})\neq Q^t(y|\mathbf{x})$.

In driver drowsiness estimation from EEG signals, EEG signals from a new subject are in the target domain, while EEG signals from the $z^{\mathrm{th}}$ existing subject are in the $z^{\mathrm{th}}$ source domain. A single data sample consists of the feature vector for a single EEG epoch in either domain. Though the features in source and target domains are extracted in the same way, generally their marginal and conditional probability distributions are different, i.e., $P^z(\mathbf{x})\neq P^t(\mathbf{x})$ and $Q^z(y|\mathbf{x})\neq Q^t(y|\mathbf{x})$, because different subjects usually have similar but distinct drowsy neural responses. As a result, data from a source domain cannot represent data in the target domain accurately, and must be integrated with some target domain data to induce the target domain regression function.

\subsection{The Learning Framework}

Because
\begin{align}
f(\mathbf{x})=Q(y|\mathbf{x})=\frac{P(\mathbf{x},y)}{P(\mathbf{x})}=\frac{Q(\mathbf{x}|y)P(y)}{P(\mathbf{x})},
\end{align}
to use the data in the $z^{\mathrm{th}}$ source domain in the target domain, we need to minimize the distance between the marginal and conditional probability distributions in the two domains by ensuring that\footnote{Strictly speaking, we should also make sure $P^z(y)$ is close to $P^t(y)$. In this paper we assume $P^z(y)$ and $P^t(y)$ are close. Our future research will consider the general case that $P^z(y)$ and $P^t(y)$ are different.} $P^z(\mathbf{x})$ is close to $P^t(\mathbf{x})$, and $Q^z(\mathbf{x}|y)$ is also close to $Q^t(\mathbf{x}|y)$.

Assume both the output and each dimension of the input vector have zero mean. Then, the regression function can be written as
\begin{align}
f(\mathbf{x})=\boldsymbol{\alpha}^T\mathbf{x} \label{eq:f}
\end{align}
where $\boldsymbol{\alpha}$ is the regression parameter vector to be found. The learning framework of OwARR is then formulated as:
\begin{align}
f=&\arg \min\limits_f\sum_{i=1}^{n}\ (y_i-f(\mathbf{x}_i))^2+w^t\sum_{i=n+1}^{n+m}(y_i-f(\mathbf{x}_i))^2 \nonumber \\
&\qquad \quad +\lambda [d(P^z,P^t)+ d(Q^z,Q^t)]- \gamma \tilde{r}^2(y,f(\mathbf{x})) \label{eq:g}
\end{align}
where $\lambda$ and $\gamma$ are non-negative regularization parameters, and $w^t$ is the overall weight for target domain samples, which should be larger than 1 so that more emphasis is given to target domain samples than source domain samples.

Briefly speaking, the first two terms in (\ref{eq:g}) minimize the sum of squared errors in the source domain and target domain, respectively. The 3rd term minimizes the distance between the marginal and conditional probability distributions in the two domains. The last term maximizes the approximate sample Pearson correlation coefficient between $y$ and $f(\mathbf{x})$, which helps avoid the undesirable situation that the regression output is (nearly) a constant.

In the next subsections we will explain how to compute the individual terms in (\ref{eq:g}).

\subsection{Sum of Squared Error Minimization}

Let
\begin{align}
X&=[\mathbf{x}_1, ...,\mathbf{x}_{n+m}]^T \label{eq:X}\\
\mathbf{y}&=[y_1,...,y_{n+m}]^T \label{eq:y}
\end{align}
where the first $n$ $\mathbf{x}_i$ and $y_i$ are the column input vectors and the corresponding outputs in the source domain, the next $m$ $\mathbf{x}_i$ and $y_i$ are the column input vectors and the corresponding outputs in the target domain, and $T$ is the matrix transpose operation.

Define $E\in R^{(n+m)\times(n+m)}$ as a diagonal matrix with
\begin{align}
E_{ii}=\left\{\begin{array}{ll}
                1, & 1\le i\le n \\
                w^t, &  n+1 \le i \le n+m
              \end{array}\right. \label{eq:E}
\end{align}
Then, the first two terms in (\ref{eq:g}) can be rewritten as
\begin{align}
&\sum_{i=1}^{n}(y_i-f(\mathbf{x}_i))^2+w^t\sum_{i=n+1}^{n+m}(y_i-f(\mathbf{x}_i))^2\nonumber \\
=&\sum_{i=1}^{n+m}E_{ii}(y_i-\boldsymbol{\alpha}^T\mathbf{x}_i)^2 \nonumber\\ =&(\mathbf{y}^T-\boldsymbol{\alpha}^TX^T)E(\mathbf{y}-X\boldsymbol{\alpha}) \label{eq:l3}
\end{align}

The optimal selection of $w^t$ is very important to the performance of the OwARR algorithm. In this paper we use
\begin{align}
w^t=\max(2,\sigma \cdot n/m) \label{eq:wt}
\end{align}
where $\sigma$ is a positive adjustable parameter, based on the following heuristics: 1) when $m$ is small, each target domain sample should have a large weight so that the target domain is not overwhelmed by the source domain; 2) as $m$ increases, the weight on the target domain samples should decrease gradually so that the source domain is not overwhelmed by the target domain; and, 3) the target domain samples should always have larger weights than the source domain samples because eventually the regression model will be applied to the target domain.

\subsection{Marginal Probability Distribution Adaptation}

As in \cite{Long2014,Quanz2009,drwuACII2015,drwuSMC2015}, we compute $d(P^z,P^t)$ using the maximum mean discrepancy (MMD):
\begin{align}
d(P^z,P^t)&=\left[\frac{1}{n}\sum_{i=1}^nf(\mathbf{x}_i)-\frac{1}{m}\sum_{i=n+1}^{n+m}f(\mathbf{x}_i)\right]^2 \nonumber \\ &=\boldsymbol{\alpha}^TXM_PX\boldsymbol{\alpha} \label{eq:DfKP}
\end{align}
where $M_P\in R^{(n+m)\times (n+m)}$ is the MMD matrix:
\begin{align}
(M_P)_{ij}=\left\{\begin{array}{ll}
                             \frac{1}{n^2},& 1\le i \le n, 1\le j \le n  \\
                             \frac{1}{m^2}, & n+1\le i \le n+m, \\
                                            &  n+1\le j \le n+m \\
                             \frac{-1}{nm}, & \text{otherwise}
                           \end{array}\right. \label{eq:M0}
\end{align}

\subsection{Conditional Probability Distribution Adaptation} \label{sect:CPDA}

In \cite{Long2014,drwuACII2015,drwuSMC2015} a classification problem is considered, and it is more straightforward to perform conditional probability distribution adaptation. In this subsection we first briefly introduce the technique used there, and then describe in detail how we can perform conditional probability distribution adaptation in regression in a similar way, with the help of FSs \cite{Zadeh1965,Klir1995}, which have been widely used in EEG feature extraction \cite{Coyle2009, Lotte2009,Khushaba2011} and pattern recognition \cite{Palaniappan2000,Lin2006,Hsu2011}.

\subsubsection{Conditional Probability Distribution Adaptation for Classification}

Let $\mathcal{D}_c^z=\{\mathbf{x}_i|\mathbf{x}_i\in \mathcal{D}^z\wedge y_i=c\}$ be the set of samples in Class $c$ ($c=1,...,C$) of the $z^{\mathrm{th}}$ source domain, $\mathcal{D}_c^t=\{\mathbf{x}_i|\mathbf{x}_i\in \mathcal{D}^t\wedge y_i=c\}$ be the set of samples in Class $c$ of the target domain, $n_c=|\mathcal{D}_c^z|$, and $m_c=|\mathcal{D}_c^t|$. Then, the distance between the conditional probability distributions in source and target domains is computed as the sum of the Euclidian distances between the class means in the two domains \cite{Long2014,drwuSMC2015,drwuACII2015}, i.e.,
\begin{align}
d(Q^z,Q^t)=\sum_{c=1}^C\left[\frac{1}{n_c}\sum\limits_{\mathbf{x}_i\in \mathcal{D}_c^z} f(\mathbf{x}_i)-\frac{1}{m_c}\sum\limits_{\mathbf{x}_i\in\mathcal{D}_c^t}f(\mathbf{x}_i)\right]^2 \label{eq:DfKQ}
\end{align}

\subsubsection{Conditional Probability Distribution Adaptation for Regression}

\begin{figure}[htpb]\centering
\subfigure[]{\label{fig:FSs}     \includegraphics[width=.48\linewidth,clip]{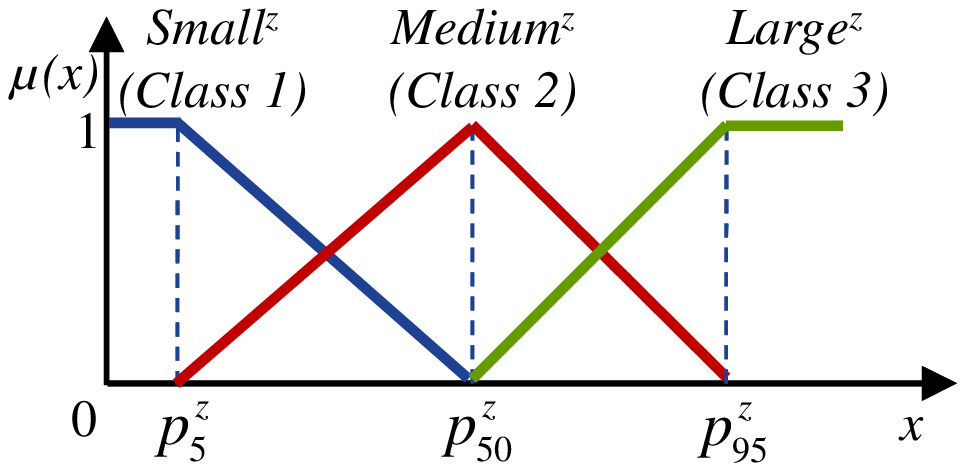}}
\subfigure[]{\label{fig:FSt}     \includegraphics[width=.48\linewidth,clip]{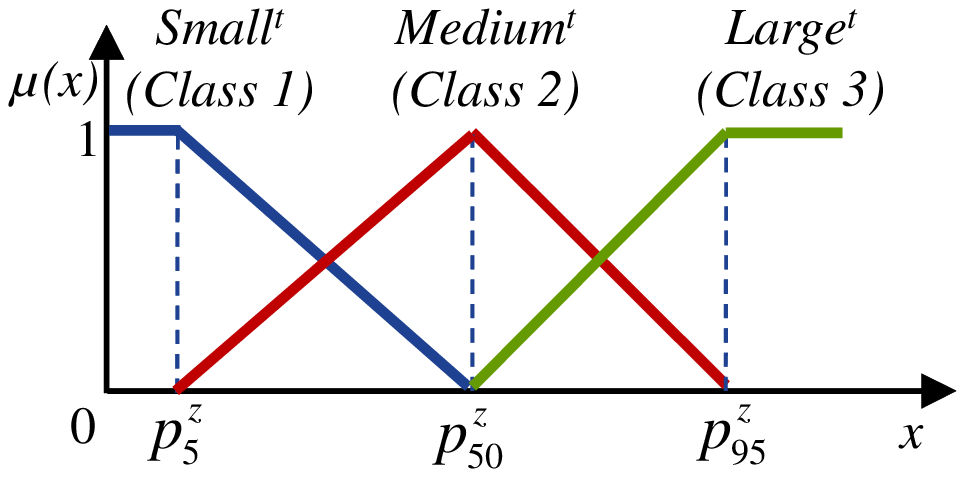}}
\caption{The three FSs in (a) the $z^{\mathrm{th}}$ source domain, and, (b) the target domain.} \label{fig:FSst}
\end{figure}

With the help of FSs (background materials are given in Appendix), we can transform the regression problem into a ``classification" problem and hence perform conditional probability distribution adaptation using (\ref{eq:DfKQ}). First, for the $n_z$ outputs, $\{y_i^z\}_{i=1,...,n}$, in the $z^{\mathrm{th}}$ source domain, we find their 5, 50 and 95 percentile\footnote{There is a popular regression analysis method called quantile regression \cite{Koenker2005} in statistics and econometrics, which estimates either the conditional median or other quantiles of the response variable. The percentiles used in this paper are found directly from the data, and they should not be confused with quantile regression.} values, $p_5^z$, $p_{50}^z$ and $p_{95}^z$, respectively, and define three triangular FSs\footnote{There can be other ways to define these FSs, e.g., we could use Gaussian FSs instead of triangular FSs, use $p_{10}^z$, $p_{50}^z$ and $p_{90}^z$ instead of $p_5^z$, $p_{50}^z$ and $p_{95}^z$, use other than three FSs in each domain, or use type-2 FSs \cite{Mendel2001} instead of type-1. Three type-1 triangular FSs are used here for simplicity. More discussions on the sensitivity of the OwARR algorithm to the number of type-1 triangular FSs are given in Section~\ref{sect:parameters}.}, $Small^z$, $Medium^z$ and $Large^z$, based on them, as shown in Fig.~\ref{fig:FSs}.  In this way, we can ``classify" the outputs in the $z^{\mathrm{th}}$ source domain into three fuzzy classes, $Small^z$, $Medium^z$ and $Large^z$, corresponding to the different classes in a traditional crisp classification problem. However, note that in the traditional crisp classification problem a sample can only belong to one class. For the fuzzy classes here, a sample can belong to more than one class simultaneously, at different degrees.

Denote Class $Small^z$ as Class 1, Class $Medium^z$ as Class 2, Class $Large^z$ as Class 3, and the membership degree of $y_i^z$ in Class $c$ as $\mu_{ic}^z$. We then normalize each $\mu_{ic}^z$ according to its class, i.e.,
\begin{align}
\bar{\mu}_{ic}^z= \displaystyle{\frac{\mu_{ic}^z}{\sum_{i=1}^n\mu_{ic}^z}},\quad  i=1,...,n; c=1,2,3
\end{align}

Similarly, we also find $p_5^t$, $p_{50}^t$ and $p_{95}^t$ from the $m$ target domain outputs  $\{y_i^t\}_{i=n+1,...,n+m}$, define three FSs, $Small^t$, $Medium^t$ and $Large^t$, as shown in Fig.~\ref{fig:FSt}, and compute the corresponding normalized $\bar{\mu}_{ic}^t$, $i=n+1,...,n+m$, $c=1,2,3$.

Finally, similar to (\ref{eq:DfKQ}), the distance between the conditional probability distributions in the target domain and the $z^{\mathrm{th}}$ source domain is computed as:
\begin{align}
d(Q^z,Q^t)=\sum_{c=1}^3\left[\sum_{\mathbf{x}_i\in \mathcal{D}^z}\bar{\mu}_{ic}^zf(\mathbf{x}_i)-\sum_{\textbf{x}_i\in \mathcal{D}^t}\bar{\mu}_{ic}^tf(\mathbf{x}_i)\right]^2 \label{eq:DfKQ2}
\end{align}
Substituting (\ref{eq:f}) into (\ref{eq:DfKQ2}), it follows that
\begin{align}
d(Q^z,Q^t)&=\sum_{c=1}^3\left[\sum_{\mathbf{x}_i\in \mathcal{D}^z}\bar{\mu}_{ic}^z\boldsymbol{\alpha}^T \mathbf{x}_i-\sum_{\mathbf{x}_i\in \mathcal{D}^t}\bar{\mu}_{ic}^t\boldsymbol{\alpha}^T\mathbf{x}_i\right]^2\nonumber \\
&=\sum_{c=1}^3 \boldsymbol{\alpha}^TXM_cX\boldsymbol{\alpha} =\boldsymbol{\alpha}^TXM_QX\boldsymbol{\alpha} \label{eq:DfKQ3}
\end{align}
where
\begin{align}
M_Q=M_1+M_2+M_3 \label{eq:MQ}
\end{align}
in which $M_1$, $M_2$ and $M_3$ are MMD matrices computed as:
\begin{align}
(M_c)_{ij}=\left\{\begin{array}{ll}
                    \bar{\mu}_{ic}^z\bar{\mu}_{jc}^z, & \mathbf{x}_i,\,\mathbf{x}_j\in \mathcal{D}_c^z \\
                    \bar{\mu}_{ic}^t\bar{\mu}_{jc}^t, &  \mathbf{x}_i,\,\mathbf{x}_j\in \mathcal{D}_c^t \\
                     -\bar{\mu}_{ic}^z\bar{\mu}_{jc}^t,& \mathbf{x}_i\in \mathcal{D}_c^z,\, \mathbf{x}_j\in \mathcal{D}_c^t \\
                     -\bar{\mu}_{ic}^t\bar{\mu}_{jc}^z,&  \mathbf{x}_i\in \mathcal{D}_c^t,\, \mathbf{x}_j\in \mathcal{D}_c^z\\
                    0, & \mbox{otherwise}
                  \end{array}\right.
\end{align}

\subsection{Maximize the Approximate Sample Pearson Correlation Coefficient}

The sample Pearson correlation coefficient $r(y,f(\mathbf{x}))$ is defined as \cite{Walpole2007}:
\begin{align}
r(y,f(\mathbf{x}))&=\frac{\mathbf{y}^TX\boldsymbol{\alpha}}{\parallel\mathbf{y}\parallel\cdot \parallel X\boldsymbol{\alpha}\parallel}\nonumber \\
 &=\frac{\mathbf{y}^TX\boldsymbol{\alpha}}{\sqrt{\mathbf{y}^T\mathbf{y}}\cdot \sqrt{\boldsymbol{\alpha}^TX^T X\boldsymbol{\alpha}}}
\end{align}
and hence
\begin{align}
r^2(y,f(\mathbf{x}))=\frac{\boldsymbol{\alpha}^TX^T\mathbf{yy}^TX\boldsymbol{\alpha}}{\mathbf{y}^T\mathbf{y}\cdot \boldsymbol{\alpha}^TX^T X\boldsymbol{\alpha}}
\end{align}

Note that $r^2(y,f(\mathbf{x}))$ has $\boldsymbol{\alpha}$ in the denominator, so it is very challenging to find a closed-form solution to maximize it. However, observe that $r^2(y,f(\mathbf{x}))$ increases as $\boldsymbol{\alpha}^TX^T\mathbf{yy}^TX\boldsymbol{\alpha}$ increases, and decreases as $\boldsymbol{\alpha}^TX^T X\boldsymbol{\alpha}$ increases. So, instead of maximizing $r^2(y,f(\mathbf{x}))$ directly, in this paper we try to maximize the following function:
\begin{align}
\tilde{r}^2(y,f(\mathbf{x}))&= \frac{\boldsymbol{\alpha}^TX^T\mathbf{yy}^TX
\boldsymbol{\alpha}-\boldsymbol{\alpha}^TX^TX\boldsymbol{\alpha}}{\mathbf{y}^T\mathbf{y}}\nonumber \\
&=\frac{\boldsymbol{\alpha}^TX^T(\mathbf{yy}^T-I)X\boldsymbol{\alpha}}{\mathbf{y}^T\mathbf{y}}\label{eq:r}
\end{align}
where $I\in R^{(n+m)\times(n+m)}$ is an identity matrix. $\tilde{r}^2(y,f(\mathbf{x}))$ has the same property as $r^2(y,f(\mathbf{x}))$, i.e., $\tilde{r}^2(y,f(\mathbf{x}))$ increases as $\boldsymbol{\alpha}^TX^T\mathbf{yy}^TX\boldsymbol{\alpha}$ increases, and decreases as $\boldsymbol{\alpha}^TX^T X\boldsymbol{\alpha}$ increases.

\subsection{The Closed-Form Solution}

Substituting (\ref{eq:l3}), (\ref{eq:DfKP}), (\ref{eq:DfKQ3}) and (\ref{eq:r}) into (\ref{eq:g}), we can rewrite it as
\begin{align}
\boldsymbol{\alpha}=&\arg \min\limits_{\boldsymbol{\alpha}}\ (\mathbf{y}^T-\boldsymbol{\alpha}^TX^T)E(\mathbf{y}-X\boldsymbol{\alpha})\nonumber \\
  &\qquad\quad +  \lambda\boldsymbol{\alpha}^TX^T(M_P+M_Q)X\boldsymbol{\alpha}\nonumber \\
  &\qquad\quad +\gamma \frac{\boldsymbol{\alpha}^TX^T(I-\mathbf{yy}^T)X\boldsymbol{\alpha}}{\mathbf{y}^T\mathbf{y}}  \label{eq:f3}
\end{align}
Setting the derivative of the objective function above to 0 leads to
\begin{align}
\boldsymbol{\alpha}=\left[X^T\left(E+\lambda M_P+ \lambda M_Q+\gamma \frac{I-\mathbf{yy}^T}{\mathbf{y}^T\mathbf{y}}\right)X\right]^{-1}X^TE\mathbf{y} \label{eq:alpha}
\end{align}

\subsection{The Complete OwARR Algorithm}

The pseudo-code for the complete OwARR algorithm is described in Algorithm~1. We first perform OwARR for each source domain separately, and then construct the final regression model as a weighted average of these base models, where the weight is the inverse of the training accuracy of the corresponding base model. The final regression model will then be applied to future unlabeled data.

\begin{algorithm}[h] 
\KwIn{$Z$ source domains, where the $z^\mathrm{th}$ ($z=1,...,Z$) domain has $n_z$ samples $\{\mathbf{x}_i^z,y_i^z\}_{i=1,...,n_z}$\;
\hspace*{10mm} $m$ target domain samples, $\{\mathbf{x}_j^t,y_j^t\}_{j=1,...,m}$\;
\hspace*{10mm} Parameters $\lambda$, $\gamma$, and $\sigma$ in (\ref{eq:wt}).}
\KwOut{The OwARR regression model.}
\For{$z=1,2,...,Z$}{
Construct $X$ in (\ref{eq:X}), $\mathbf{y}$ in (\ref{eq:y}), $E$ in (\ref{eq:E}), $M_P$ in (\ref{eq:M0}), and $M_Q$ in (\ref{eq:MQ})\;
Compute $\boldsymbol{\alpha}$ by (\ref{eq:alpha}) and record it as $\boldsymbol{\alpha}^z$\;
Use $\boldsymbol{\alpha}^z$ to estimate the outputs for the $n_z+m$ samples from both domains and record the root mean squared error as $a^z$\;
Assign the $z^{\mathrm{th}}$ regression model a weight $w^z=1/a^z$\;}
\textbf{Return} $f(\mathbf{x})=\frac{\sum_{z=1}^Zw^z(\boldsymbol{\alpha}^z)^T\mathbf{x}}{\sum_{z=1}^Zw^z}$.
\caption{The OwARR algorithm.} \label{alg:OwARR}
\end{algorithm}

\section{OwARR-SDS} \label{sect:OwARR-SDS}

A SDS procedure for online classification problems has been proposed in \cite{drwuACII2015}. In this paper it is extended to regression problems by using the fuzzy classes again defined in Fig.~\ref{fig:FSst}. The primary goal of SDS is to reduce the computational cost of OwARR, because when there is a large number of source domains, performing OwARR for each source domain and then aggregating the base models would be very time-consuming.

Assume there are $Z$ different source domains. For the $z^\mathrm{th}$ source domain, we first compute $\mathbf{m}_c^z$ ($c=1,2,3$), the mean vector of each fuzzy class. Then, we also compute $\mathbf{m}_c^t$, the mean vector of each fuzzy class in the target domain, from the $m$ labeled samples. The distance between the two domains is:
\begin{align}
d(z,t)=\sum_{c=1}^3 ||\mathbf{m}_c^z-\mathbf{m}_c^t|| \label{eq:dST}
\end{align}
We next cluster the $Z$ numbers, $\{d(z,t)\}_{z=1,...,Z}$, by $k$-means clustering, and finally choose the cluster that has the smallest centroid, i.e., the source domains that are closest to the target domain. In this way, on average we only need to perform OwARR for $Z/k$ source domains. We used $k=2$ in this paper.

The pseudo-code for the complete OwARR-SDS algorithm is described in Algorithm~2. We first use SDS to select the $Z'$ closest source domains, and then perform DA for each selected source domain separately. The final regression model is a weighted average of these base models, with the weight being the inverse of the training accuracy of the corresponding base model.

\begin{algorithm}[htpb] 
\KwIn{$Z$ source domains, where the $z^\mathrm{th}$ ($z=1,...,Z$) domain has $n_z$ samples $\{\mathbf{x}_i^z,y_i^z\}_{i=1,...,n_z}$\;
\hspace*{10mm} $m$ target domain samples, $\{\mathbf{x}_j^t,y_j^t\}_{j=1,...,m}$\;
\hspace*{10mm} $\lambda$, $\gamma$, $\sigma$ in (\ref{eq:wt}), and $k$ in $k$-means clustering.}
\KwOut{The OwARR-SDS regression model.}
\tcp{SDS starts}
\eIf{$m==0$}{
Select all $Z$ source domains\;
Go to OwARR.}
{\For{$z=1,2,...,Z$}{
Compute $d(z,t)$, the distance between the target domain and the $z^\mathrm{th}$ source domain, by (\ref{eq:dST}). }
Cluster $\{d(z,t)\}_{z=1,...,Z}$ by $k$-means clustering; \\
Select the $Z'$ source domains that belong to the cluster with the smallest centroid. }
\tcp{SDS ends; OwARR starts}
\For{$z=1,2,...,Z'$}{
Construct $X$ in (\ref{eq:X}), $\mathbf{y}$ in (\ref{eq:y}), $E$ in (\ref{eq:E}), $M_P$ in (\ref{eq:M0}), and $M_Q$ in (\ref{eq:MQ})\;
Compute $\boldsymbol{\alpha}$ by (\ref{eq:alpha}) and record it as $\boldsymbol{\alpha}^z$\;
Use $\boldsymbol{\alpha}^z$ to estimate the outputs for the $n_z+m$ samples from both domains and record the root mean squared error as $a^z$\;
Assign the $z^{\mathrm{th}}$ regression model a weight $w^z=1/a^z$\;}
\tcp{OwARR ends}
\textbf{Return} $f(\mathbf{x})=\frac{\sum_{z=1}^{Z'}w^z(\boldsymbol{\alpha}^z)^T\mathbf{x}}{\sum_{z=1}^{Z'}w^z}$.
\caption{The OwARR-SDS algorithm.} \label{alg:OwARR-SDS}
\end{algorithm}

\section{Experiments and Discussions} \label{sect:experiments}

Experimental results on driver drowsiness estimation from EEG signals are presented in this section to demonstrate the performance of OwARR and OwARR-SDS.

\subsection{Experiment Setup}

We reused the experiment setup and data in \cite{drwuaBCI2015}. 16 healthy subjects with normal or corrected to normal vision were recruited to participate in a sustained-attention driving experiment \cite{Chuang2012,Chuang2014}, which consisted of a real vehicle mounted on a motion platform with 6 degrees of freedom immersed in a 360-degree virtual-reality scene. Each participant read and signed an informed consent form before the experiment began. Each experiment lasted for about 60-90 minutes and was conducted in the afternoon when the circadian rhythm of sleepiness reached its peak. To induce drowsiness during driving, the virtual-reality scenes simulated monotonous driving at a fixed 100 km/h speed on a straight and empty highway. During the experiment, lane-departure events were randomly applied every 5-10 seconds, and participants were instructed to steer the vehicle to compensate for these perturbations as quickly as possible. Subjects' cognitive states and driving performance were monitored via a surveillance video camera and the vehicle trajectory throughout the experiment. The response time in response to the perturbation was recorded and later converted to drowsiness index. Meanwhile, participants' scalp EEG signals were recorded using a 32-channel (30-channel EEGs plus 2-channel earlobes) 500 Hz Neuroscan NuAmps Express system (Compumedics Ltd., VIC, Australia).

The Institutional Review Board of the Taipei Veterans General Hospital approved the experimental protocol.

\subsection{Evaluation Process and Performance Measures} \label{sect:process}

The complete procedure for the application of OwARR for driver drowsiness estimation is shown in Algorithm~3. Compared with Algorithm~1 on OwARR for a generic application, here we also include detailed EEG data pre-processing and feature extraction steps. Observe that although the feature extraction methods for different auxiliary subjects have the same steps, their parameters (channels removed, principal components used, ranges used in normalization) may be different, so we need to record them for each auxiliary subject so that the features in the individual regression models can be computed correctly.

\begin{algorithm}[htpb] 
\KwIn{EEG data and the corresponding RTs\;
\hspace*{10mm} $\lambda$, $\gamma$, and $\sigma$ in OwARR (Algorithm~2)\;
\hspace*{10mm} $t_{\max}$, the duration of the calibration period\;
\hspace*{10mm} $\Delta t$, the time interval (in second) between two successive epochs in calibration.}
\KwOut{The OwARR regression model.}
Set $t=0$, and start the calibration\;
\While{$t<t_{\max}$}{
\If{$t\ge 30$}{
Compute the drowsiness index $y$ in (\ref{eq:y})\;
\tcp{Pre-processing}
Extract 30s EEG data in $[t-30,\ t]$\;
Band-pass filter the EEG signals to $[0,\ 50]$ Hz\;
Down-sample to 250Hz\;
Re-reference to averaged earlobes\;
Compute the PSD in the $[4,\ 7.5]$ Hz theta band for each channel\;
Convert the PSDs to dB for each channel\;}
Wait until $t=t+\Delta t$\;}
\tcp{OwARR}
\For{$z=1,2,...,Z$}{
\tcp{Feature extraction}
Concatenate each channel of the powers of the $z^\mathrm{th}$ subject with the corresponding powers of the new subject\;
Remove channels which have at least one power larger than a certain threshold\;
Normalize the powers of each remaining channel to mean 0 and std 1\;
Put all the powers in a matrix, whose rows represent different EEG channels\;
Extract a few leading principal components of the power matrix that account for 95\% of the variance\;
Find the corresponding scores of the leading principal components\;
Normalize each dimension of the scores to $[0,\,1]$\;
Collect the scores for each epoch as features\;
Record the parameters of the feature extraction method (channels removed, principal components used, ranges used in normalization) as $FE^z$\;
\tcp{OwARR training}
Construct $X$ in (\ref{eq:X}), $\mathbf{y}$ in (\ref{eq:y}), $E$ in (\ref{eq:E}), $M_P$ in (\ref{eq:M0}), and $M_Q$ in (\ref{eq:MQ})\;
Compute $\boldsymbol{\alpha}$ by (\ref{eq:alpha}) and record it as $\boldsymbol{\alpha}^z$\;
Use $\boldsymbol{\alpha}^z$ to estimate the outputs for all known samples and record the root mean squared error as $a^z$\;
Assign the $z^{\mathrm{th}}$ regression model a weight $w^z=1/a^z$\;}
\textbf{Return} The OwARR regression model $f(\mathbf{x})=\frac{\sum_{z=1}^Zw^z(\boldsymbol{\alpha}^z)^T\mathbf{x}^z}{\sum_{z=1}^Zw^z}$, where $\mathbf{x}^z$ is the feature vector extracted using $FE^z$.
\caption{OwARR for driver drowsiness estimation.} \label{alg:Online}
\end{algorithm}

From the experiments we already knew the drowsiness indices for all $\sim$1200 epochs. To evaluate the performances of different algorithms, for each subject, we used up to 100 epochs in a randomly chosen continuous block for calibration, and the rest $\sim$1100 epochs for testing. Every time when five epochs were acquired, we computed the testing performance to show how the performance of the regression models changed over time. We ran this evaluation process 30 times, each time with a randomly chosen 100-epoch calibration block, to obtain statistically meaningful results. Finally, we repeated this entire process 15 times so that each subject had a chance to be the ``15th" subject.

The primary performance measured used in this paper is the root mean squared error (RMSE) between the $\sim$1100 true drowsiness indices and the corresponding estimates for the testing epochs, which is optimized in the object functions of all algorithms. The secondary performance measure is the correlation coefficient (CC) between the true drowsiness indices and the estimates.

\subsection{Preprocessing and Feature Extraction}

The 16 subjects had different lengths of experiment, because the disturbances were presented randomly every 5-10 seconds. Data from one subject was not correctly recorded, so we used only 15 subjects. To ensure fair comparison, we used only the first 3,600 seconds data for each subject.

\begin{figure}[htpb]\centering
\includegraphics[width=\linewidth,clip]{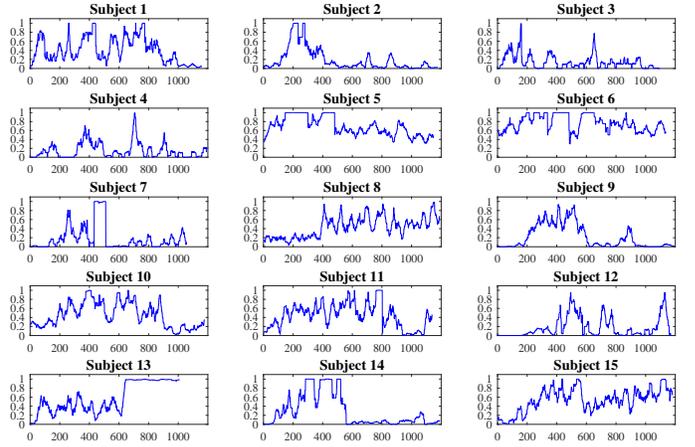} \caption{Drowsiness indices of the 15 subjects.} \label{fig:resTime}
\end{figure}

We defined a function \cite{Wei2015,drwuaBCI2015} to map the response time $\tau$ to a drowsiness index $y\in[0, 1]$:
\begin{align}
y=\max\left\{0,\,\frac{1-e^{-(\tau-\tau_0)}}{1+e^{-(\tau-\tau_0)}}\right\}
\end{align}
$\tau_0=1$ was used in this paper, as in \cite{drwuaBCI2015}. The drowsiness indices were then smoothed using a 90-second square moving-average window to reduce variations. This does not reduce the sensitivity of the drowsiness index because the cycle lengths of drowsiness fluctuations are longer than 4 minutes \cite{Makeig1993}. The smoothed drowsiness indices for the 15 subjects are shown in Fig.~\ref{fig:resTime}. Observe that each subject had some drowsiness indices at or close to 1, indicating drowsy driving.

We used EEGLAB \cite{Delorme2004} for EEG signal preprocessing. A band-pass filter (1-50 Hz) was applied to remove high-frequency muscle artifacts, line-noise contamination and DC drift. Next the EEG data were downsampled from 500 Hz to 250 Hz and re-referenced to averaged earlobes.

\begin{figure}[tb]\centering
\subfigure[]{\label{fig:drowy1}     \includegraphics[width=.48\linewidth,clip]{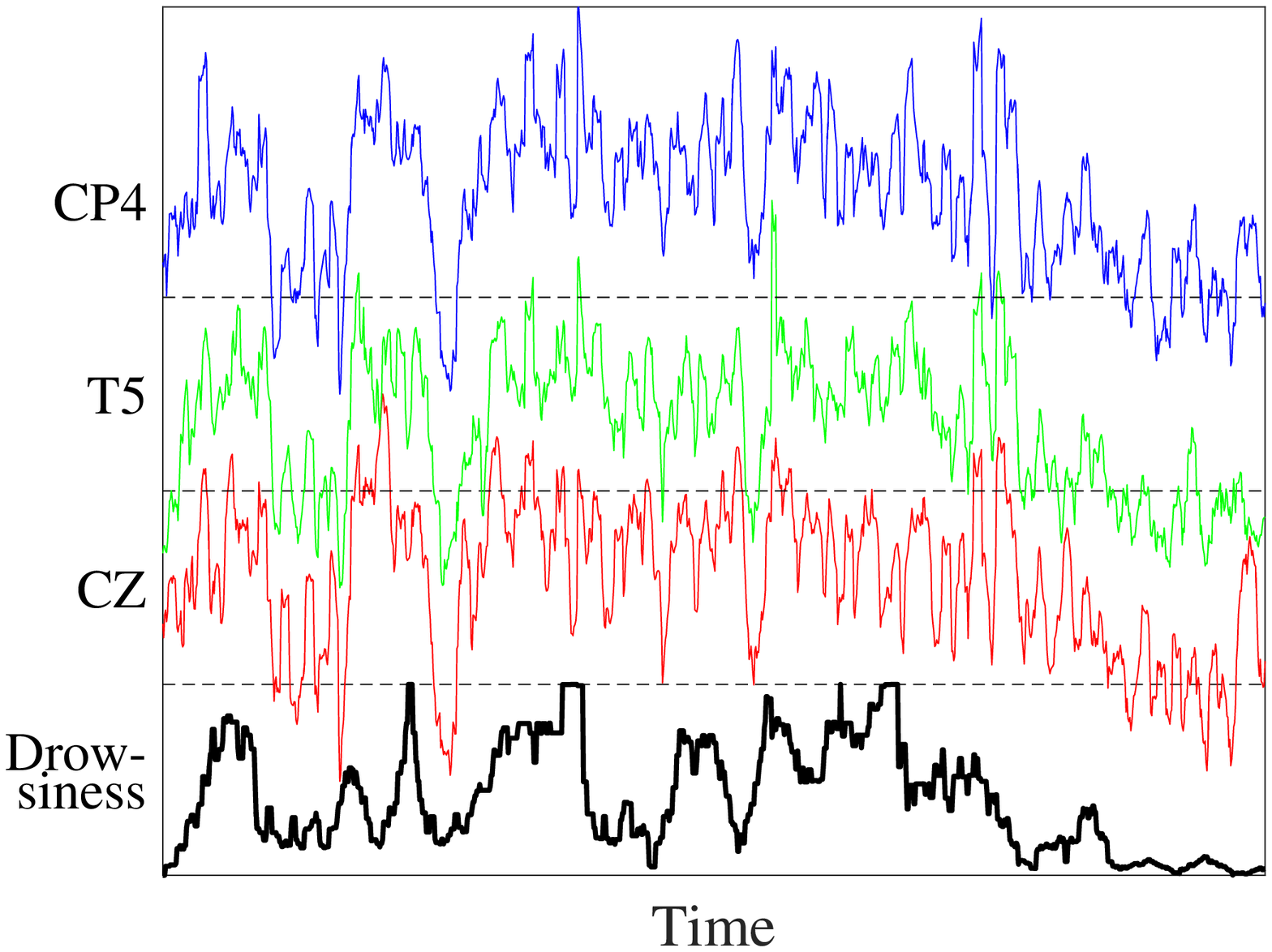}}
\subfigure[]{\label{fig:drowy2}     \includegraphics[width=.48\linewidth,clip]{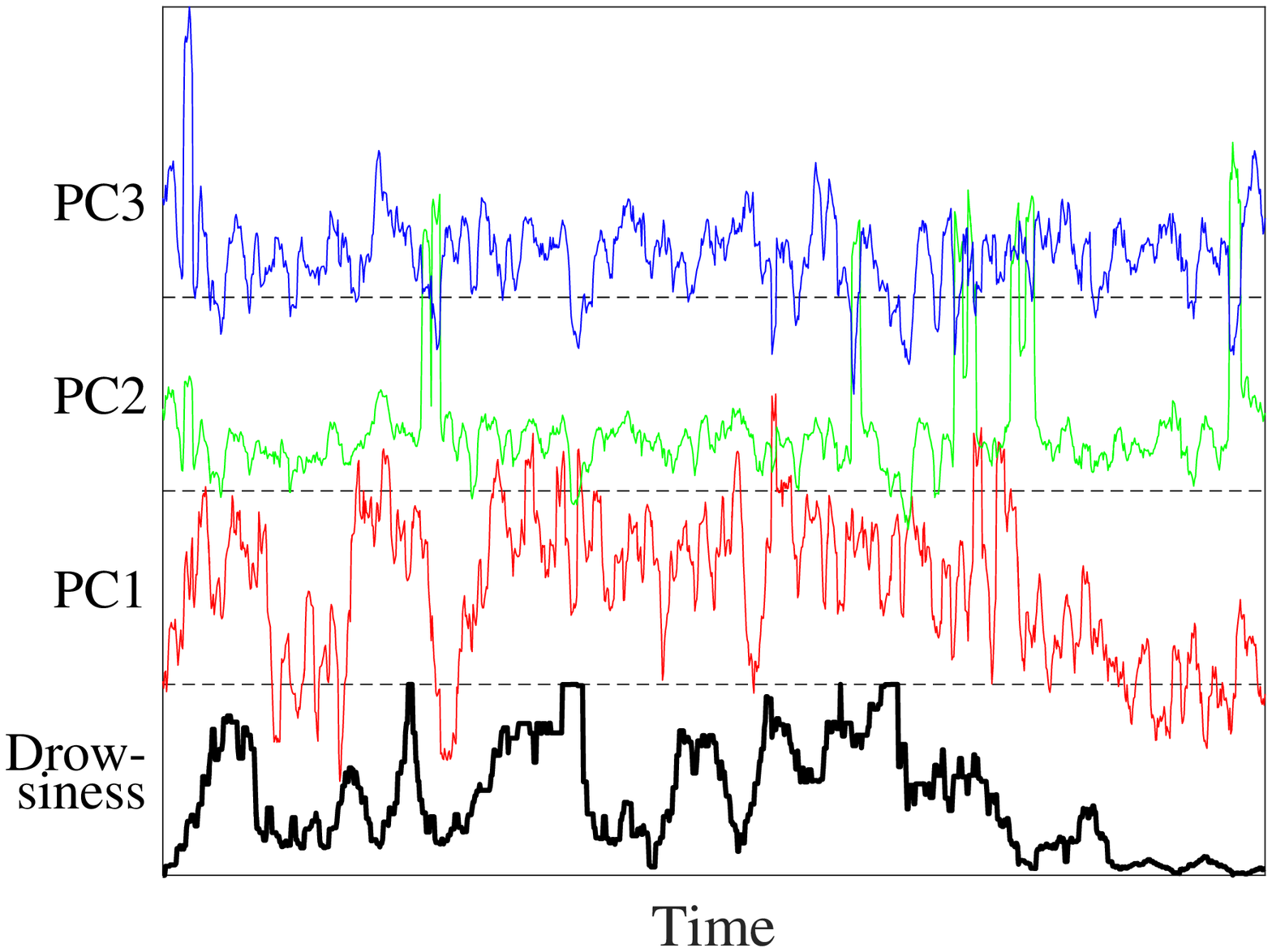}}
\caption{EEG features and the corresponding drowsiness indices for Subject 1. (a) Theta band powers for three selected channels; (b) The top three principal component (PC) features.}
\end{figure}

We tried to predict the drowsiness index for each subject every three seconds. All 30 EEG channels were used in feature extraction. We epoched 30-second EEG signals right before each sample point, and computed the average power spectral density (PSD) in the theta band (4-7.5 Hz) for each channel using Welch's method \cite{Welch1967}, as research \cite{Makeig1996} has shown that theta band spectrum is a strong indicator of drowsiness. The theta band powers for three selected channels and the corresponding drowsiness index for Subject 1 are shown in Fig.~\ref{fig:drowy1}. Observe that drowsiness index has strong correlations with the theta band powers.

Next, we converted the 30 theta band powers to dBs. To remove noises or bad channel readings, we removed channels whose maximum dBs were larger than 20. We then normalized the dBs of each remaining channel to mean zero and standard deviation one, and extracted a few (usually around 10) leading principal components, which accounted for 95\% of the variance. The projections of the theta band powers onto these principal components were then normalized to $[0, 1]$ and used as our features. Three such features for Subject 1 are shown in Fig.~\ref{fig:drowy2}. Observe that the score on the first principal component has obvious correlation with the drowsiness index, suggesting that estimating the drowsiness index from the scores on the principal components is possible.

\subsection{Algorithms} \label{sect:algorithms}

We compared the performances of OwARR and OwARR-SDS with three other algorithms introduced in \cite{drwuaBCI2015}:
\begin{enumerate}
\item Baseline 1 (BL1), which combines data from all 14 existing subjects, builds a ridge regression model \cite{Hastie2009}, and applies it to the new subject. That is, BL1 tries to build a subject-independent regression model and ignores data from the new subject completely.
\item Baseline 2 (BL2), which builds a ridge regression model using only subject-specific calibration samples from the new subject. That is, BL2 ignores data from existing subjects completely.
\item DAMF, which builds 14 ridge regression models by combining data from each auxiliary subject with data from the new subject, respectively, and then uses a weighted average to obtain the final regression model. The weights are also the inverse of the training RMSEs, as in Algorithms~1 and 2.
\end{enumerate}
The ridge parameter $\sigma=0.01$ was used in the above three algorithms, as in \cite{drwuaBCI2015}. For OwARR and OwARR-SDS, we used $\sigma=0.2$, $\lambda=10$, and $\gamma=0.5$. However, as will be shown in Section~\ref{sect:parameters}, OwARR and OwARR-SDS are robust to these three parameters.

\subsection{Regression Performance Comparison} \label{sect:results}

\begin{figure*}[htpb]\centering
\subfigure[]{\label{fig:RMSEs}     \includegraphics[width=.7\linewidth,clip]{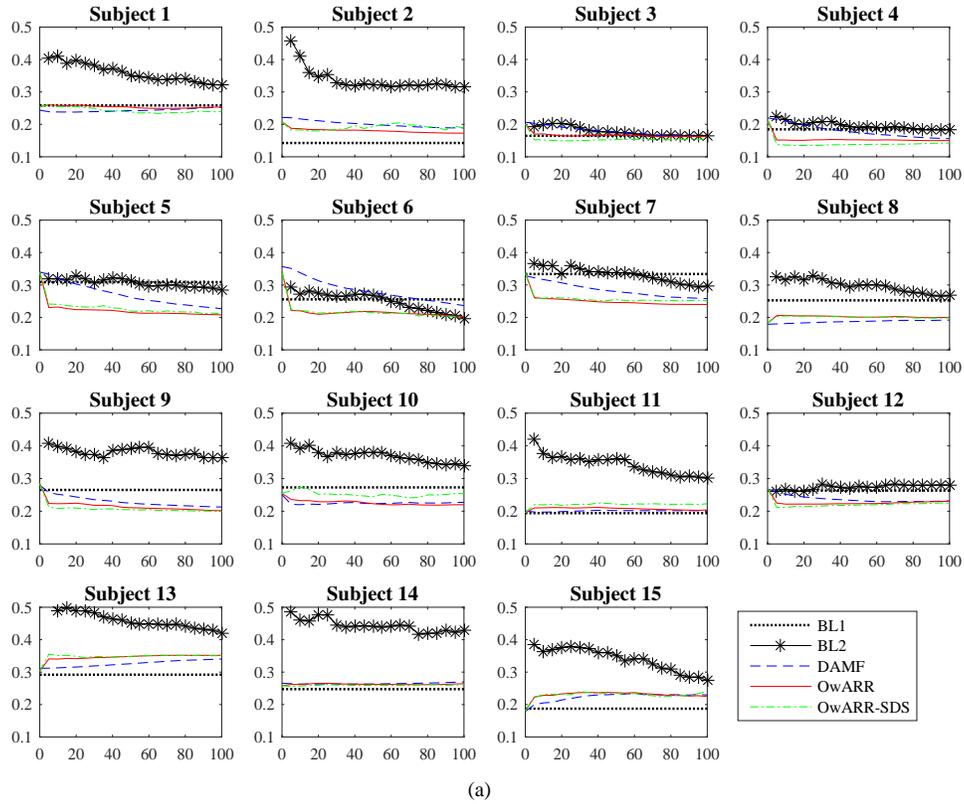}}
\subfigure[]{\label{fig:CCs}     \includegraphics[width=.7\linewidth,clip]{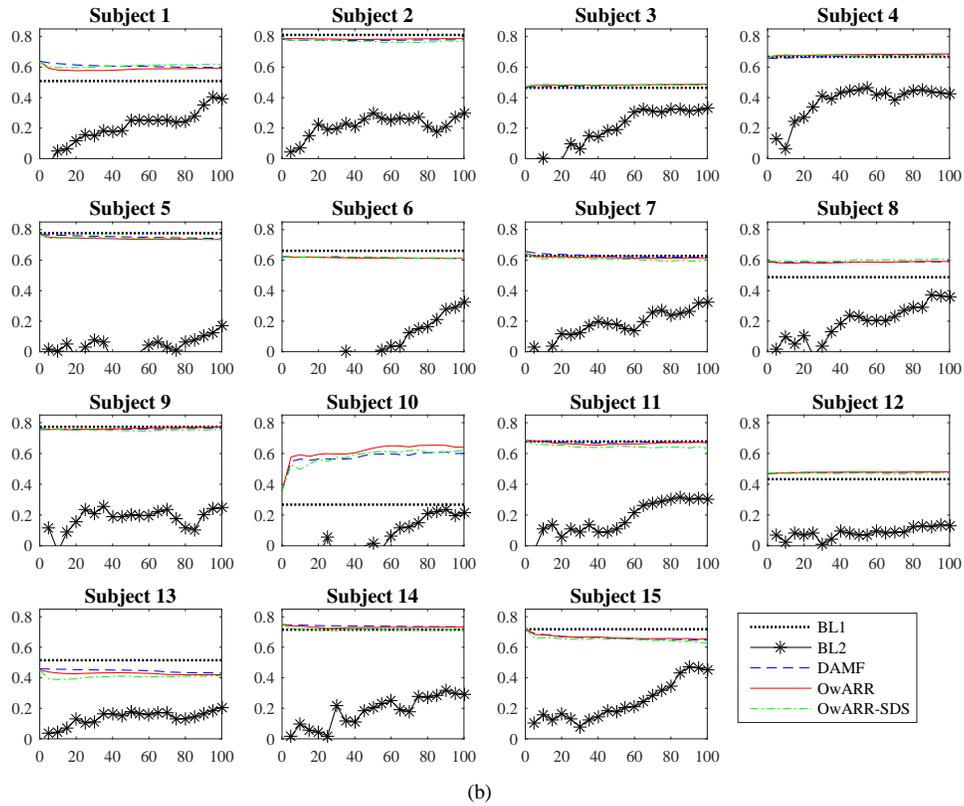}}
\caption{Average performances of the five algorithms for each individual subject. Horizontal axis: $m$, the number of subject-specific calibration samples. (a) RMSE; (b) CC.} \label{fig:ind}
\end{figure*}

The average RMSEs and CCs for the five algorithms across the 15 subjects are shown in Fig.~\ref{fig:mean}, and the RMSEs and CCs for the individual subjects are shown in Fig.~\ref{fig:ind}. Observe that DAMF, OwARR and OwARR-SDS had very similar CCs, and all of them were better than the CCs of BL1 and BL2. Additionally:

\begin{figure}[htpb]\centering
\subfigure[]{\label{fig:mRMSE}     \includegraphics[width=.48\linewidth,clip]{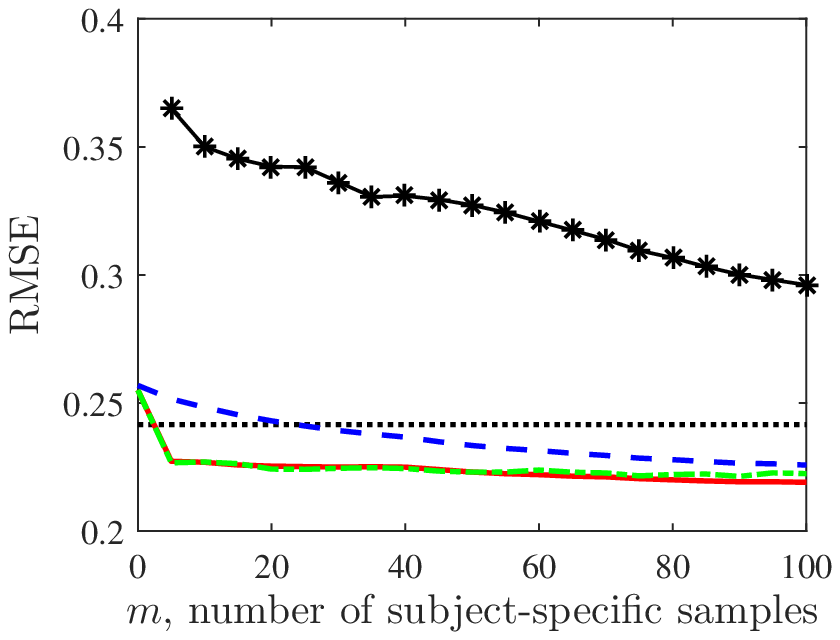}}
\subfigure[]{\label{fig:mCC}     \includegraphics[width=.48\linewidth,clip]{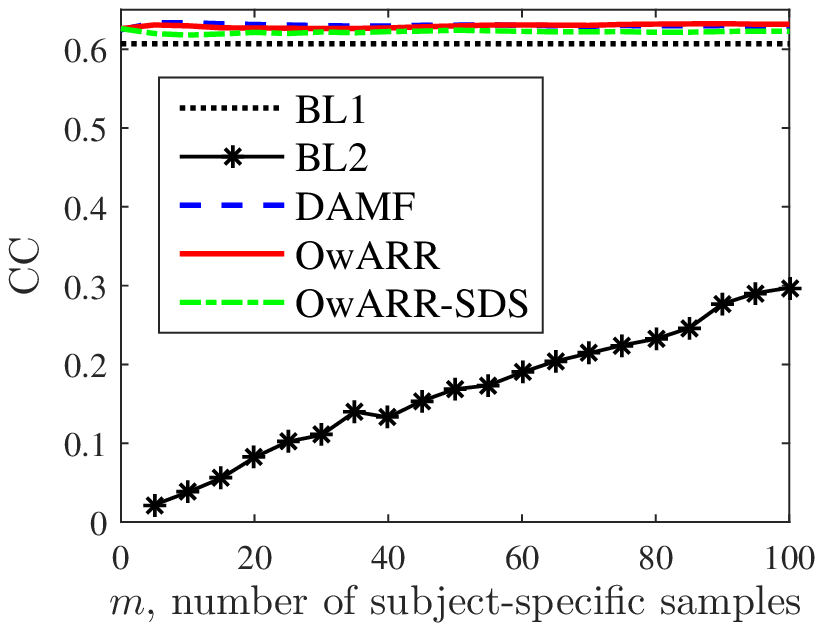}}
\caption{Average performances of the five algorithms across the 15 subjects. (a) RMSE; (b) CC.} \label{fig:mean}
\end{figure}

\begin{enumerate}
\item Except for BL1, whose model does not depend on $m$, all the other four algorithms gave smaller RMSEs as $m$ increased, which is intuitive.
\item BL1 had the smallest RMSE when $m=0$. However, as $m$ increased, DAMF, OwARR and OwARR-SDS quickly outperformed BL1. This suggests that there is large individual difference among the subjects, and hence a subject-independent model is not desirable.
\item Because BL2 used only subject-specific calibration data, it cannot build a model when $m=0$, i.e., when there was no subject-specific calibration data at all. However, all the other four methods can, because they can make use of data from other subjects. BL2's performance was the worst, because it cannot get enough training when there is only a small number of subject-specific calibration samples.
\item OwARR and OwARR-SDS had almost identical average RMSEs, which were smaller than those of BL1, BL2 and DAMF. More importantly, the RMSEs of OwARR and OwARR-SDS almost converged as soon as the first batch of subject-specific samples were added, suggesting that they only need very few subject-specific samples to train, which is very desirable in practical calibration.
\end{enumerate}
In summary, the three DA based approaches generally had better performance than BL1, which does not use subject-specific data at all, and also BL2, which does not use auxiliary data at all. This suggests that DA is indeed beneficial. Moreover, our proposed OwARR and OwARR-SDS achieved the best overall performances among the five algorithms.

We also performed two-way Analysis of Variance (ANOVA) for each $m$ to check if the RMSE differences among the five algorithms were statistically significant, by setting the subjects as a random effect. Two-way ANOVA showed statistically significant differences among them ($p<0.01$) for all $m$. Then, non-parametric multiple comparison tests using Dunn's procedure \cite{Dunn1961,Dunn1964} were used to determine if the difference between any pair of algorithms was statistically significant, with a $p$-value correction using the False Discovery Rate method \cite{Benjamini1995}. The $p$-values are shown in Table~\ref{tab:Dunn}, where the statistically significant ones are marked in bold. Observe that the differences between OwARR and the other three algorithms (BL1, BL2, and DAMF) were always statistically significant when $m>0$, so were the differences between OwARR-SDS and the two baseline algorithms. The difference between OwARR-SDS and DAMF was statistically significant for $m\in[5, 75]$. There was no statistically significant difference between OwARR and OwARR-SDS.

\begin{table*}[ht] \centering \setlength{\tabcolsep}{4mm}
\caption{$p$-values of non-parametric multiple comparisons.}   \label{tab:Dunn}
\begin{tabular}{c|ccccccc}   \hline
 & OwARR  & OwARR  & OwARR  & OwARR-SDS  & OwARR-SDS  & OwARR-SDS  & OwARR-SDS  \\
$m$ &  vs BL1 &  vs BL2 &  vs DAMF &  vs BL1 &  vs BL2 &  vs DAMF &  vs OwARR \\ \hline
0   & \textbf{.0007}  &       N/A        &  .4073         &   \textbf{.0011} &     N/A        &.5091            &   .5000\\
5   & \textbf{.0000}  &  \textbf{.0000}  & \textbf{.0000}  &  \textbf{.0000} & \textbf{.0000} & \textbf{.0000}  &  .4813\\
10  & \textbf{.0000}  &  \textbf{.0000}  & \textbf{.0000}  &  \textbf{.0000} & \textbf{.0000} & \textbf{.0000}  &  .3888\\
15  & \textbf{.0000}  &  \textbf{.0000}  & \textbf{.0000}  &  \textbf{.0000} & \textbf{.0000} & \textbf{.0000}  &  .4075\\
20  & \textbf{.0000}  &  \textbf{.0000}  & \textbf{.0000}  &  \textbf{.0000} & \textbf{.0000} & \textbf{.0000}  &   .4795\\
25  & \textbf{.0000}  &  \textbf{.0000}  & \textbf{.0000}  &  \textbf{.0000} & \textbf{.0000} & \textbf{.0000}  &   .4658\\
30  & \textbf{.0000}  &  \textbf{.0000}  & \textbf{.0000}  &  \textbf{.0000} & \textbf{.0000} & \textbf{.0000}  &   .4153\\
35  & \textbf{.0000}  &  \textbf{.0000}  & \textbf{.0001}  &  \textbf{.0000} & \textbf{.0000} & \textbf{.0001}  &   .4364\\
40  & \textbf{.0000}  &  \textbf{.0000}  & \textbf{.0002}  &  \textbf{.0000} & \textbf{.0000} & \textbf{.0004}  &   .3956\\
45  & \textbf{.0000}  &  \textbf{.0000}  & \textbf{.0004}  &  \textbf{.0000} & \textbf{.0000} & \textbf{.0006}  &   .4378\\
50  & \textbf{.0000}  &  \textbf{.0000}  & \textbf{.0005}  &  \textbf{.0000} & \textbf{.0000} & \textbf{.0013}  &   .3766\\
55  & \textbf{.0000}  &  \textbf{.0000}  & \textbf{.0007}  &  \textbf{.0000} & \textbf{.0000} & \textbf{.0035}  &   .2898\\
60  & \textbf{.0000}  &  \textbf{.0000}  & \textbf{.0011}  &  \textbf{.0000} & \textbf{.0000} & \textbf{.0108}  &   .2132\\
65  & \textbf{.0000}  &  \textbf{.0000}  & \textbf{.0015}  &  \textbf{.0000} & \textbf{.0000} & \textbf{.0097}  &   .2573\\
70  & \textbf{.0000}  &  \textbf{.0000}  & \textbf{.0024}  &  \textbf{.0000} & \textbf{.0000} & \textbf{.0143}  &   .2527\\
75  & \textbf{.0000}  &  \textbf{.0000}  & \textbf{.0038}  &  \textbf{.0000} & \textbf{.0000} & \textbf{.0133}  &   .3122\\
80  & \textbf{.0000}  &  \textbf{.0000}  & \textbf{.0047}  &  \textbf{.0000} & \textbf{.0000} &          .0294  &   .2304\\
85  & \textbf{.0000}  &  \textbf{.0000}  & \textbf{.0058}  &  \textbf{.0000} & \textbf{.0000} &          .0513  &   .1785\\
90  & \textbf{.0000}  &  \textbf{.0000}  & \textbf{.0071}  &  \textbf{.0000} & \textbf{.0000} &          .0383  &   .2387\\
95  & \textbf{.0000}  &  \textbf{.0000}  & \textbf{.0086}  &  \textbf{.0000} & \textbf{.0000} &          .1091  &   .1205\\
100 & \textbf{.0000}  &  \textbf{.0000}  & \textbf{.0117}  &  \textbf{.0000} & \textbf{.0000} &          .1179  &   .1352\\  \hline
\end{tabular}
\end{table*}

Finally, we can conclude that given the same amount of subject-specific calibration data, OwARR and OwARR-SDS can achieve significantly better estimation performance than the other three approaches. Or, in other words, given a desired RMSE, OwARR and OwARR-SDS require significantly less subject-specific calibration data than the other three approaches. For example, in Fig.~\ref{fig:mRMSE}, the average RMSE for BL2 when $m=100$ was $0.2988$, whereas OwARR and OwARR-SDS can achieve even smaller RMSEs without using any subject-specific calibration samples. The average RMSEs for OwARR and OwARR-SDS when $m=5$ were $0.2347$ and $0.2348$, respectively, whereas DAMF needed at least 45 subject-specific samples to achieve these RMSEs, and BL2 needed at least 100 samples.

\subsection{Computational Cost}

In this subsection we compare the computational cost of the five algorithms, particularly, OwARR and OwARR-SDS, because the primary goal of SDS is to down-select the number of auxiliary subjects and hence to reduce the computational cost of OwARR.

\begin{figure}[htpb]\centering
\includegraphics[width=\linewidth]{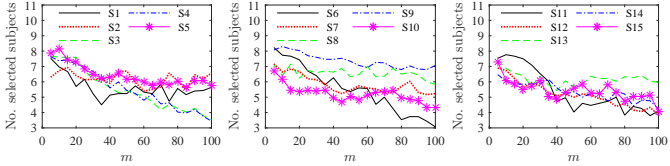} \caption{Average number of similar subjects selected by SDS.} \label{fig:SDS}
\end{figure}

Fig.~\ref{fig:SDS} shows the average number of similar subjects selected by SDS for the 15 subjects. Observe that most of the time fewer than seven subjects (half of the number of auxiliary subjects) were selected.

\begin{figure}[htpb]\centering
\includegraphics[width=.8\linewidth]{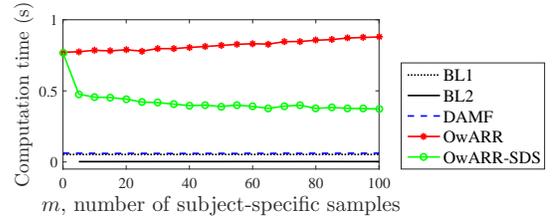} \caption{Average training time of the five algorithms. Note that BL1 overlaps with DAMF.} \label{fig:compCost}
\end{figure}

To quantify the computational cost of the five algorithms, we show in Fig.~\ref{fig:compCost} the training times for different $m$, averaged over 10 runs and across the 15 subjects. The platform was a Dell XPS 13 notebook, with Intel Core i7-5500M CPU@2.40GHz, 8GB memory, and 256GB solid state drive. The software was Matlab R2015b running in 64-bit Windows 10 Pro. Each algorithm was optimized to the best ability of the authors. Observe that the training time of BL1, BL2, and DAMF was almost constant, whereas the training time of OwARR increased monotonically as $m$ increased. Interestingly, the training time of OwARR-SDS decreased slightly as $m$ increased, because Fig.~\ref{fig:SDS} shows that generally the average number of similar subjects selected by SDS decreased as $m$ increased.

The computational costs of OwARR and OwARR-SDS were much higher than DAMF, because they used more sophisticated DA approaches. However, except for $m=0$, at which point OwARR and OwARR-SDS had identical training time, the training time of OwARR-SDS was on average only about 49\% of OwARR. This 51\% computation time saving is very worthwhile when the number of source domains is very large and hence computing OwARR for all the source domains is too slow.

\begin{figure}[tb]\centering
\subfigure[]{\label{fig:nDomains}     \includegraphics[width=.48\linewidth,clip]{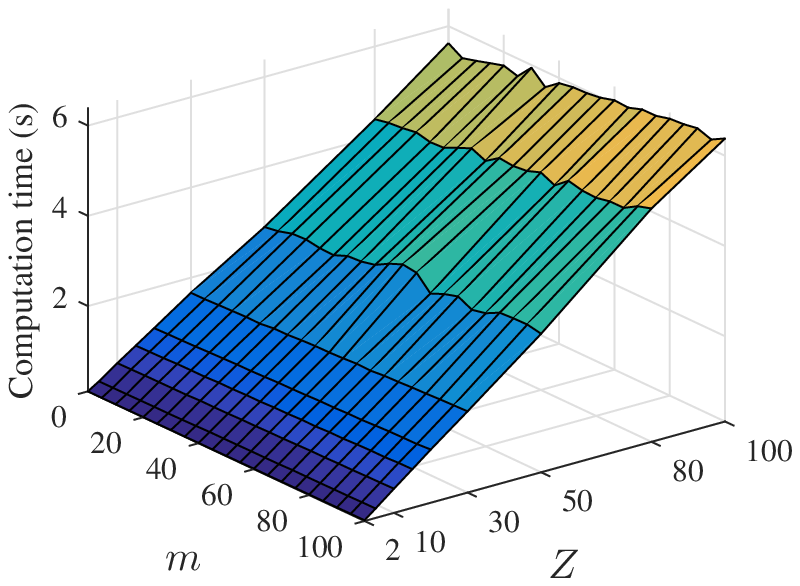}}
\subfigure[]{\label{fig:nSamples}     \includegraphics[width=.48\linewidth,clip]{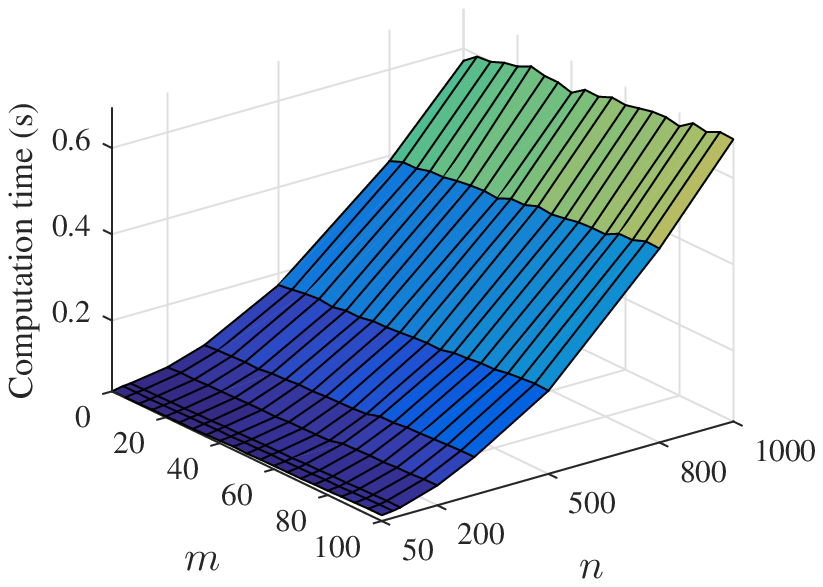}}
\caption{Scalability of OwARR with respect to (a) the number of source domains (each domain had about 1,200 samples); (b) the number of samples in each source domain (14 source domains were used).} \label{fig:scalability}
\end{figure}

We also investigated the scalability of OwARR with respect to $Z$, the number of source domains, and $n$, the number of samples in each source domain. Because we only had 14 source domains in this dataset, we bootstrapped them to create additional domains when $Z\ge 14$. The results are shown in Fig.~\ref{fig:scalability}. Observe from Fig.~\ref{fig:nDomains} that the computational cost of OwARR increased linearly with the number of source domains, which is intuitive, because OwARR performs DA for each source domain separately and then aggregates the results. However, Fig.~\ref{fig:nSamples} shows that the computational cost of OwARR increased superlinearly with the number of samples in the source domains. Least-squares curve fitting found that the computation time was about $0.0000021\cdot n^{1.8}+0.035$ seconds, i.e., the computational cost is $O(n^{1.8})$ for 14 source domains.

Finally, it is important to note that the above analyses are only for the training of the algorithms. Once the training is done, the resulting OwARR and OwARR-SDS models can be executed much faster.

\subsection{Robustness to Noises}

It is also important to study the robustness of the five algorithms to noises. According to \cite{Zhu2004}, there are two types of noises: \emph{class noise}, which is the noise on the model outputs, and \emph{attribute noise}, which is the noise on the model inputs. In this subsection we focus on the attribute noise.

\begin{figure}[htpb]\centering
\includegraphics[clip,width=.8\linewidth]{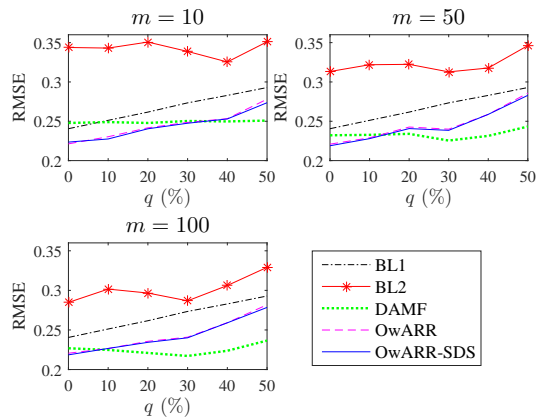} \caption{Average RMSEs of the five algorithms with respect to different attribute noise levels.} \label{fig:noises}
\end{figure}

As in \cite{Zhu2004}, for each model input, we randomly replaced $q\%$ ($q=0,\,10,...,50$) of all epochs from the new subject with a uniform noise between its minimum and maximum values. After this was done for both the training and testing data, we trained the five algorithms on the corrupted training data and then tested their performances on the corrupted testing data. The RMSEs for three different $m$ (the number of labeled subject-specific samples), averaged across 15 subjects with five runs per subject, are shown in Fig.~\ref{fig:noises}. Observe that as the noise level $q$ increased, generally all algorithms had worse RMSEs. OwARR and OwARR-SDS still had the smallest RMSEs among the five when $q$ was small. However, when $q$ increased, DAMF became the best. This suggests that OwARR and OwARR-SDS may not be as robust as DAMF with respect to attribute noises, but when the noise level is low, the performance improvement achieved from the sophisticated optimizations in OwARR and OwARR-SDS dominates, and hence they are still the best algorithms among the five. When the noise level is high, we may need some noise handling approaches, e.g., noise correction \cite{Zhu2004}, before applying OwARR and OwARR-SDS.

\subsection{Parameter Sensitivity Analysis} \label{sect:parameters}

The OwARR algorithm has three adjustable parameters: $\sigma$, which determines the weight $w^t$ for the target domain samples; $\lambda$, which is a regularization parameter minimizing the distances between the marginal and conditional probability distributions in the source and target domains; and $\gamma$, which maximizes the approximate Pearson correlation coefficient between the true and estimated outputs. It is interesting to study whether all of them are necessary.

\begin{figure}[htpb]\centering
\includegraphics[width=.8\linewidth,clip]{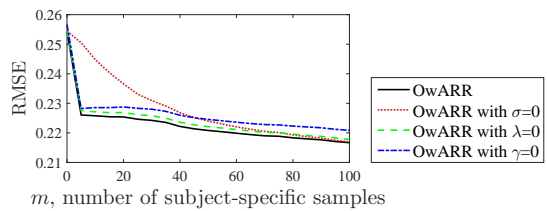} \caption{Average RMSEs of OwARR when different regularization terms are removed.} \label{fig:components}
\end{figure}

For this purpose, we constructed three modified versions of the OwARR algorithms by setting $\sigma$, $\lambda$ and $\gamma$ to zero, respectively, and compared their average RMSEs with that of the original OwARR. The results are shown in Fig.~\ref{fig:components}. Observe that the original OwARR had better performance than all three modified versions, suggesting that all three parameters in OwARR contributed to its superior performance.

\begin{figure}[tb]\centering
\subfigure[]{\label{fig:wt}     \includegraphics[width=.48\linewidth,clip]{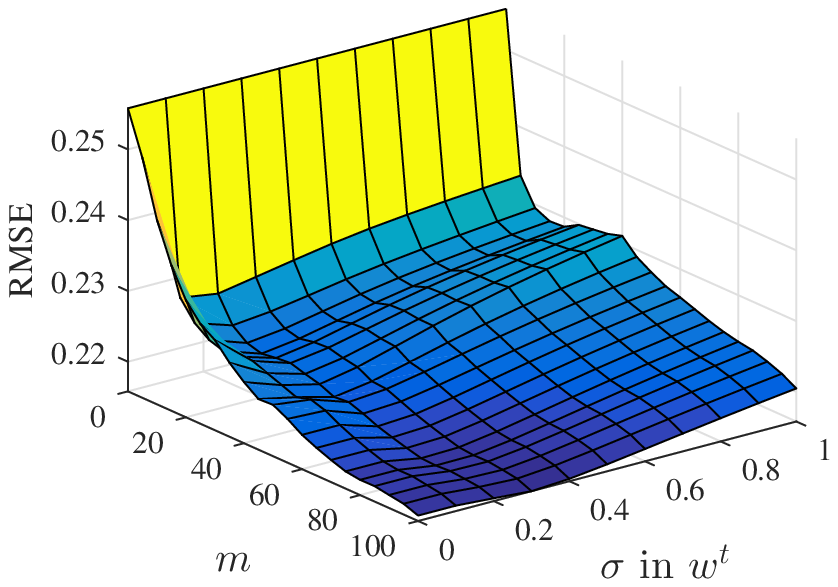}}
\subfigure[]{\label{fig:lambda}     \includegraphics[width=.48\linewidth,clip]{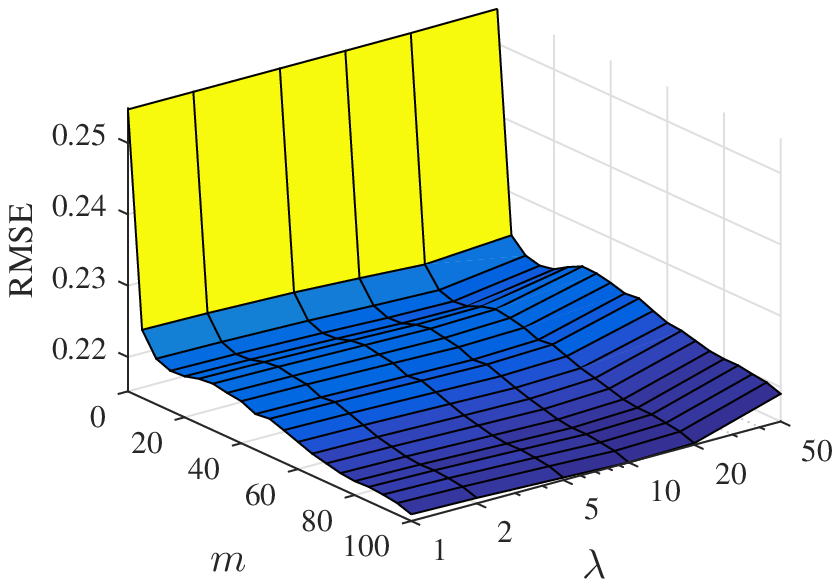}}
\subfigure[]{\label{fig:gamma}     \includegraphics[width=.48\linewidth,clip]{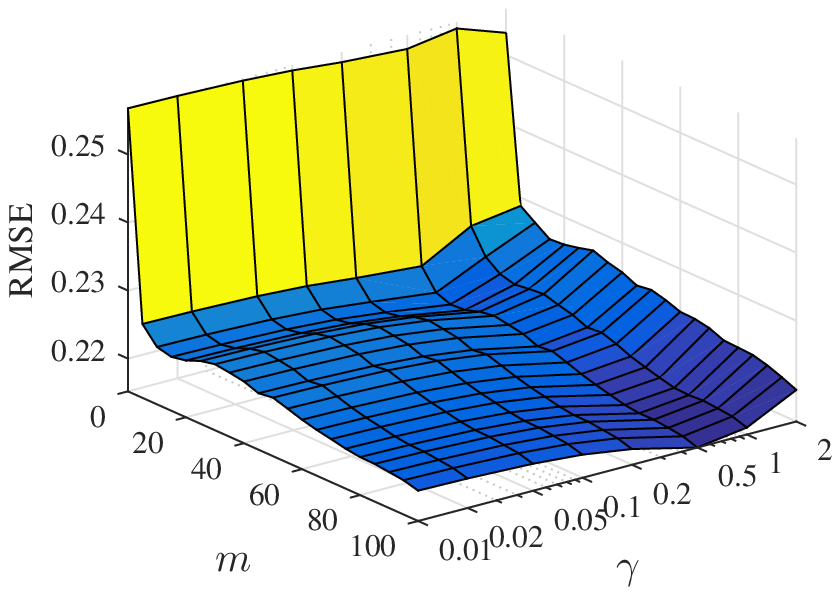}}
\subfigure[]{\label{fig:nFSs}     \includegraphics[width=.48\linewidth,clip]{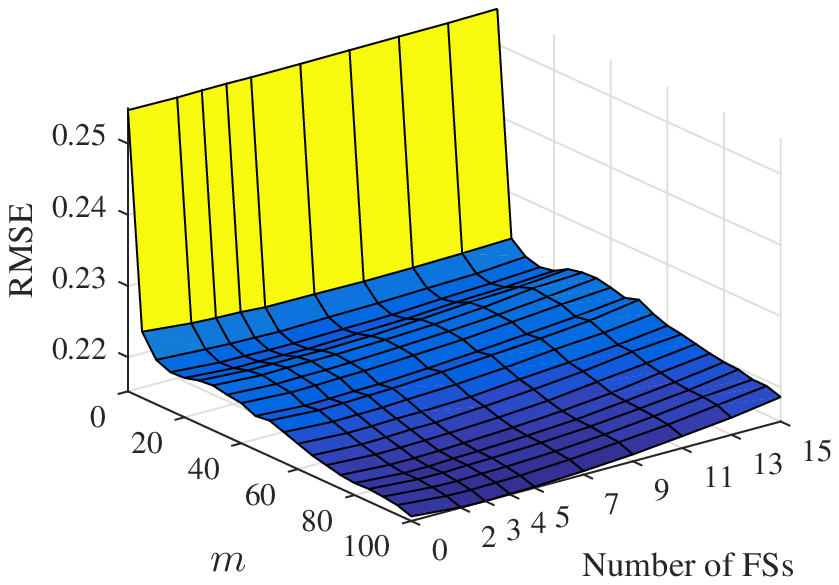}}
\caption{Average RMSEs of OwARR across the 15 subjects for different parameter values. (a) $\sigma$ in $w^t$; (b) $\lambda$; (c) $\gamma$; (d) number of FSs in conditional probability distribution adaptation.} \label{fig:para}
\end{figure}

Next we studied the sensitivity of OwARR to the three adjustable parameters, $\sigma$, $\lambda$ and $\gamma$. The results are shown in Fig.~\ref{fig:para}(a)-(c). Observe that OwARR is robust to $\sigma$ in the range of $[0.1, 0.4]$, to $\lambda$ in the range of $[1, 20]$, and to $\gamma$ in the range of $[0.01, 1]$.

Additionally, three type-1 triangular FSs have been used in conditional probability distribution adaptation (Section~\ref{sect:CPDA}) in this paper for simplicity. It is also interesting to study the sensitivity of the OwARR algorithm to the number of FSs. The results are shown in Fig.~\ref{fig:nFSs}. Observe that OwARR gives the optimal performance when the number of FSs is between 2 and 5, but its performance gradually deteriorates when the number of FSs further increases. This is intuitive, because the target domain has a limited number of labeled training samples, so as the number of FSs increases, the number of target domain samples that fall into each fuzzy class decreases, and hence the computed fuzzy class means are less reliable. As a result, the distance between the conditional probability distributions [see (\ref{eq:DfKQ3})] cannot be reliably computed.

Another interesting questions is: what would be the performance of OwARR if no FSs are used at all, i.e., conditional probability distribution adaptation is disabled? This is corresponding to the left-most slice in Fig.~\ref{fig:nFSs}, where the number of FSs is zero. Observe that this results in worse RMSEs than the case that two to five FSs are used in conditional probability distribution adaptation, suggesting the FS approach is beneficial.

\subsection{Effectiveness of the Ensemble Fusion Strategy}

\begin{figure}[htpb]\centering
\includegraphics[clip,width=\linewidth]{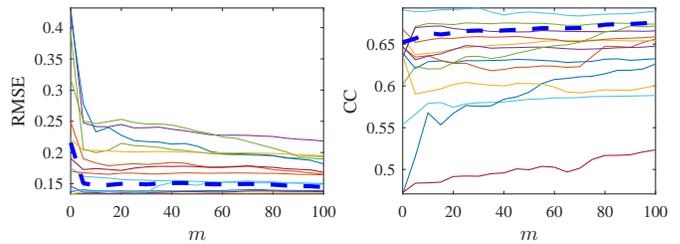} \caption{Performances of the 14 base DA models (solid curves) and the final regression model (dashed blue curve) for a typical subject.} \label{fig:ensemble}
\end{figure}

From Algorithm~1 (Algorithm~2) it is clear that the final step of OwARR (OwARR-SDS) uses ensemble learning: the base DA models are aggregated using a weighted average to obtain the final regression model, and the weight is inversely proportional to the training RMSE of the corresponding base DA model. In this subsection we study whether this fusion strategy is effective. The performances of the 14 base DA models and the final aggregation model for a typical subject are shown in Fig.~\ref{fig:ensemble}. Observe that the aggregated model is better than most base DA models, and is also close to the best base DA model (which is unknown in practice), suggesting that the fusion strategy is effective. However, it may be possible that a better fusion strategy can make the final model outperform all base DA models. This will be one of our future research directions.

\section{Conclusions and Future Research} \label{sect:conclusions}

Transfer learning, which improves learning performance in a new task by leveraging data or knowledge from other relevant tasks, represents a promising solution for handling individual differences in BCI. Previously we have proposed a weighted adaptation regularization (wAR) algorithm \cite{drwuSMC2015,drwuTNSRE2016} for offline BCI classification problems, an online weighted adaptation regularization (OwAR) algorithm \cite{drwuACII2015} for online BCI classification problems, and a SDS approach \cite{drwuACII2015,drwuSMC2015} to reduce the computational cost of wAR and OwAR. In this paper we have proposed an OwARR algorithm to extend the OwAR algorithm from classification to regression, and validated its performance on online estimation of driver drowsiness from EEG signals. Meanwhile, we have also extended the SDS algorithm for classification in \cite{drwuACII2015} to regression problems, and verified that OwARR-SDS can achieve similar performance to OwARR, but save about half of the computation time. Both OwARR and OwARR-SDS use fuzzy sets to perform part of the adaptation regularization, and OwARR-SDS also uses fuzzy sets to select the closest source domains.

Though OwARR and OwARR-SDS have demonstrated outstanding performance, they can be enhanced in a number of ways, which will be considered in our future research. First, Fig.~\ref{fig:RMSEs} shows that OwARR and OwARR-SDS had worse RMSEs than BL1 for some subjects. This indicates that they still have room for improvement: we could develop a mechanism to switch between BL1 and OwARR (OwARR-SDS) so that a more appropriate method is chosen according to the characteristics of the new subject, similar to the idea of selective TL \cite{Wei2015}. Second, we will extend OwARR and OwARR-SDS to offline calibration, where the goal is to automatically label some initially unlabeled subject-specific samples with a small number of queries \cite{drwuSMC2015}. Semi-supervised learning can be used here to enhance the learning performance. Third, in this paper we combine the base learners using a simple weighted average, where the weights of the base learners are inversely proportional to their corresponding training RMSEs. This may not be optimal because what really matter here are the testing RMSEs. In online calibration it is not easy to estimate the testing RMSEs because we do not know what samples will be encountered in the future; however, in offline calibration we can better estimate the testing performances of the base learners using a spectral meta-learner approach \cite{drwuSMLR2016}, and hence a better model fusion strategy could be developed. Fourth, similar to offline classification problems \cite{drwuSMC2014,drwuTNSRE2016,drwuRSVP2016}, in offline regression problems we can also integrate DA with active learning \cite{Settles2009,drwuEBMAL2016} to further reduce the offline calibration effort. Finally, we will apply the online and offline DA algorithms to other regression problems in BCI and beyond to cope with individual differences, e.g., estimating the continuous values of arousal, valence and dominance from speech signals \cite{drwuICME2010} in affective computing.

\section*{Acknowledgement}

Research was sponsored by the U.S. Army Research Laboratory and was accomplished under Cooperative Agreement Numbers W911NF-10-2-0022 and W911NF-10-D-0002/TO 0023. The views and the conclusions contained in this document are those of the authors and should not be interpreted as representing the official policies, either expressed or implied, of the U.S. Army Research Laboratory or the U.S Government. This work was also partially supported by the Australian Research Council (ARC) under discovery grant DP150101645.

\appendix[Fuzzy Sets (FSs)]

FS theory was first introduced by Zadeh \cite{Zadeh1965} in 1965 and has been successfully used in many areas, including modeling and control \cite{Wang1997,drwuEAAI2006}, data mining \cite{Pedrycz1998,Yager1996,drwuLS2011}, time-series prediction \cite{Versaci2003,Kasabov2002}, decision making \cite{Mendel2001,drwuBook,Ragin2000}, etc.

\begin{figure}[htpb]
\centering \includegraphics[width=.7\linewidth,clip]{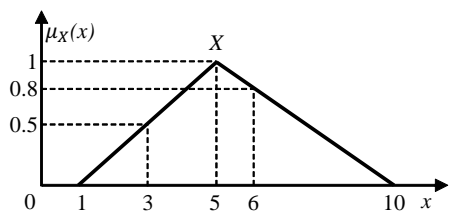} \caption{An example of a FS. } \label{fig:T1FS}
\end{figure}

A FS $X$ is comprised of a \emph{universe of discourse} $D_X$ of real numbers together with a \emph{membership function} (MF) $\mu_{_X}:\ D_X \to [0,1]$, i.e.,
\begin{align}
X=\int_{D_X}\mu_{_X}(x)/x
\end{align}
Here $\int$ denotes the collection of all points $x\in D_X$ with associated \emph{membership degree} $\mu_{_X}(x)$. An example of a FS is shown in Fig.~\ref{fig:T1FS}. The membership degrees are $\mu_X(1)=0$, $\mu_X(3)=0.5$, $\mu_X(5)=1$, $\mu_X(6)=0.8$, and $\mu_X(10)=0$. Observe that this is different from traditional (binary) sets, where each element can only belong to a set completely (i.e., with membership degree 1), or does not belong to it at all (i.e., with membership degree 0); there is nothing in between (i.e., with membership degree 0.5).

FSs are frequently used in modeling concepts in natural language, which may not have clear boundary. For example, we may define a \emph{hot} day as temperature equal to or above $30^{\circ}$C, but is $29^{\circ}$C hot? If we represent \emph{hot} as a binary set $\{x|x\ge 30\}$, then $29^{\circ}$C is not hot, because it does not belong to the binary set \emph{hot}. However, this does not completely agree with people's intuition: $29^{\circ}$C is very close to $30^{\circ}$C, and hence it is somewhat hot. If we represent \emph{hot} as a FS, we may say $29^{\circ}$C is hot with a membership degree of $0.9$, which sounds more reasonable.

\begin{thebibliography}{10}
\providecommand{\url}[1]{#1}

\bibitem{Bamdadian2013}
A.~Bamdadian, C.~Guan, K.~K. Ang, and J.~Xu, ``Improving session-to-session
  transfer performance of motor imagery-based {BCI} using adaptive extreme
  learning machine,'' in \emph{Proc. 35th Annual Int'l Conf. of the {IEEE}
  Engineering in Medicine and Biology Society ({EMBC})}, Osaka, Japan, July
  2013, pp. 2188--2191.

\bibitem{Benjamini1995}
Y.~Benjamini and Y.~Hochberg, ``Controlling the false discovery rate: A
  practical and powerful approach to multiple testing,'' \emph{Journal of the
  Royal Statistical Society, Series B (Methodological)}, vol.~57, pp. 289--300,
  1995.

\bibitem{Chuang2014}
C.-H. Chuang, L.-W. Ko, T.-P. Jung, and C.-T. Lin, ``Kinesthesia in a
  sustained-attention driving task,'' \emph{Neuroimage}, vol.~91, pp. 187--202,
  2014.

\bibitem{Chuang2012}
S.-W. Chuang, L.-W. Ko, Y.-P. Lin, R.-S. Huang, T.-P. Jung, and C.-T. Lin,
  ``Co-modulatory spectral changes in independent brain processes are
  correlated with task performance,'' \emph{Neuroimage}, vol.~62, pp.
  1469--1477, 2012.

\bibitem{Coyle2009}
D.~Coyle, G.~Prasad, and T.~McGinnity, ``Faster self-organizing fuzzy neural
  network training and a hyperparameter analysis for a brain-computer
  interface,'' \emph{{IEEE} Trans. on Systems, Man, and Cybernetics, Part {B}:
  Cybernetics}, vol.~39, no.~6, pp. 1458--1471, 2009.

\bibitem{Delorme2004}
A.~Delorme and S.~Makeig, ``{EEGLAB}: an open source toolbox for analysis of
  single-trial {EEG} dynamics including independent component analysis,''
  \emph{Journal of Neuroscience Methods}, vol. 134, pp. 9--21, 2004.

\bibitem{Devlaminck2011}
D.~Devlaminck, B.~Wyns, M.~Grosse-Wentrup, G.~Otte, and P.~Santens,
  ``Multisubject learning for common spatial patterns in motor-imagery {BCI},''
  \emph{Computational intelligence and neuroscience}, vol.~20, no.~8, 2011.

\bibitem{Dunn1961}
O.~Dunn, ``Multiple comparisons among means,'' \emph{Journal of the American
  Statistical Association}, vol.~56, pp. 62--64, 1961.

\bibitem{Dunn1964}
O.~Dunn, ``Multiple comparisons using rank sums,'' \emph{Technometrics},
  vol.~6, pp. 214--252, 1964.

\bibitem{Hastie2009}
T.~Hastie, R.~Tibshirani, and J.~Friedman, \emph{The Elements of Statistical
  Learning}.\hskip 1em plus 0.5em minus 0.4em\relax Springer, 2009.

\bibitem{Hsu2011}
W.-Y. Hsu, C.-Y. Lin, W.-F. Kuo, M.~Liou, Y.-N. Sun, A.~C. hsin Tsai, H.-J.
  Hsu, P.-H. Chen, and I.-R. Chen, ``Unsupervised fuzzy c-means clustering for
  motor imagery {EEG} recognition,'' \emph{Int'l Journal of Innovative
  Computing, Information and Control}, vol.~7, no.~8, pp. 4965--4976, 2011.

\bibitem{Jayaram2016}
V.~Jayaram, M.~Alamgir, Y.~Altun, B.~Scholkopf, and M.~Grosse-Wentrup,
  ``Transfer learning in brain-computer interfaces,'' \emph{{IEEE}
  Computational Intelligence Magazine}, vol.~11, no.~1, pp. 20--31, 2016.

\bibitem{Kang2009}
H.~Kang, Y.~Nam, and S.~Choi, ``Composite common spatial pattern for
  subject-to-subject transfer,'' \emph{Signal Processing Letters}, vol.~16,
  no.~8, pp. 683--686, 2009.

\bibitem{Kasabov2002}
N.~K. Kasabov and Q.~Song, ``{DENFIS}: {D}ynamic evolving neural-fuzzy
  inference system and its application for time-series prediction,''
  \emph{{IEEE} Trans. on Fuzzy Systems}, vol.~10, no.~2, pp. 144--154, 2002.

\bibitem{Khushaba2011}
R.~Khushaba, S.~Kodagoda, S.~Lal, and G.~Dissanayake, ``Driver drowsiness
  classification using fuzzy wavelet-packet-based feature-extraction
  algorithm,'' \emph{{IEEE} Trans. on Biomedical Engineering}, vol.~58, no.~1,
  pp. 121--131, 2011.

\bibitem{Klir1995}
G.~J. Klir and B.~Yuan, \emph{Fuzzy Sets and Fuzzy Logic: {T}heory and
  Applications}.\hskip 1em plus 0.5em minus 0.4em\relax Upper Saddle River, NJ:
  Prentice-Hall, 1995.

\bibitem{Koenker2005}
R.~Koenker, \emph{Quantile Regression}.\hskip 1em plus 0.5em minus 0.4em\relax
  Cambridge University Press, 2005.

\bibitem{Lance2012}
B.~J. Lance, S.~E. Kerick, A.~J. Ries, K.~S. Oie, and K.~McDowell,
  ``Brain-computer interface technologies in the coming decades,'' \emph{Proc.
  of the {IEEE}}, vol. 100, no.~3, pp. 1585--1599, 2012.

\bibitem{Li2010}
Y.~Li, H.~Kambara, Y.~Koike, and M.~Sugiyama, ``Application of covariate shift
  adaptation techniques in brain-computer interfaces,'' \emph{{IEEE} Trans. on
  Biomedical Engineering}, vol.~57, no.~6, pp. 1318--1324, 2010.

\bibitem{Li2009}
Y.~Li, Y.~Koike, and M.~Sugiyama, ``A framework of adaptive brain computer
  interfaces,'' in \emph{Proc. 2nd {IEEE} Int'l Conf. on Biomedical Engineering
  and Informatics ({BMEI})}, Tianjin, China, October 2009.

\bibitem{Liao2012}
L.-D. Liao, C.-T. Lin, K.~McDowell, A.~Wickenden, K.~Gramann, T.-P. Jung, L.-W.
  Ko, and J.-Y. Chang, ``Biosensor technologies for augmented brain-computer
  interfaces in the next decades,'' \emph{Proc. of the {IEEE}}, vol. 100,
  no.~2, pp. 1553--1566, 2012.

\bibitem{Lin2006}
C.-T. Lin, L.-W. Ko, I.-F. Chung, T.-Y. Huang, Y.-C. Chen, T.-P. Jung, and
  S.-F. Liang, ``Adaptive {EEG}-based alertness estimation system by using
  {ICA}-based fuzzy neural networks,'' \emph{{IEEE} Trans. on Circuits and
  Systems-I}, vol.~53, no.~11, pp. 2469--2476, 2006.

\bibitem{Long2014}
M.~Long, J.~Wang, G.~Ding, S.~J. Pan, and P.~S. Yu, ``Adaptation
  regularization: A general framework for transfer learning,'' \emph{{IEEE}
  Trans. on Knowledge and Data Engineering}, vol.~26, no.~5, pp. 1076--1089,
  2014.

\bibitem{Lotte2009}
F.~Lotte, A.~Lecuyer, and B.~Arnaldi, ``{FuRIA}: An inverse solution based
  feature extraction algorithm using fuzzy set theory for brain-computer
  interfaces,'' \emph{{IEEE} Trans. on Signal Processing}, vol.~57, no.~8, pp.
  3253--3263, 2009.

\bibitem{Lotte2015}
F.~Lotte, ``Signal processing approaches to minimize or suppress calibration
  time in oscillatory activity-based brain-computer interfaces,'' \emph{Proc.
  of the {IEEE}}, vol. 103, no.~6, pp. 871--890, 2015.

\bibitem{Makeig1993}
S.~Makeig and M.~Inlow, ``Lapses in alertness: Coherence of fluctuations in
  performance and {EEG} spectrum,'' \emph{Electroencephalography and Clinical
  Neurophysiology}, vol.~86, pp. 23--35, 1993.

\bibitem{Makeig1996}
S.~Makeig and T.~P. Jung, ``Tonic, phasic and transient {EEG} correlates of
  auditory awareness in drowsiness,'' \emph{Cognitive Brain Research}, vol.~4,
  pp. 12--25, 1996.

\bibitem{Makeig2012}
S.~Makeig, C.~Kothe, T.~Mullen, N.~Bigdely-Shamlo, Z.~Zhang, and
  K.~Kreutz-Delgado, ``Evolving signal processing for brain-computer
  interfaces,'' \emph{Proc. of the {IEEE}}, vol. 100, no. Special Centennial
  Issue, pp. 1567--1584, 2012.

\bibitem{drwuRSVP2016}
A.~Marathe, V.~Lawhern, D.~Wu, D.~Slayback, and B.~Lance, ``Improved neural
  signal classification in a rapid serial visual presentation task using active
  learning,'' \emph{{IEEE} Trans. on Neural Systems and Rehabilitation
  Engineering}, vol.~24, no.~3, pp. 333--343, 2016.

\bibitem{Mendel2001}
J.~M. Mendel, \emph{Uncertain Rule-Based Fuzzy Logic Systems: {I}ntroduction
  and New Directions}.\hskip 1em plus 0.5em minus 0.4em\relax Upper Saddle
  River, NJ: Prentice-Hall, 2001.

\bibitem{drwuBook}
J.~M. Mendel and D.~Wu, \emph{Perceptual Computing: {A}iding People in Making
  Subjective Judgments}.\hskip 1em plus 0.5em minus 0.4em\relax Hoboken, NJ:
  Wiley-{IEEE} Press, 2010.

\bibitem{Muhl2014}
C.~Muhl, B.~Allison, A.~Nijholt, and G.~Chanel, ``A survey of affective brain
  computer interfaces: principles, state-of-the-art, and challenges,''
  \emph{Brain-Computer Interfaces}, vol.~1, no.~2, pp. 66--84, 2014.

\bibitem{Palaniappan2000}
R.~Palaniappan, P.~Raveendran, S.~Nishida, and N.~Saiwaki, ``Evolutionary fuzzy
  {ARTMAP} for autoregressive model order selection and classification of {EEG}
  signals,'' in \emph{{IEEE} Int'l Conf. on Systems, Man, and Cybernetics},
  vol.~5, Nashville, TN, October 2000, pp. 3682--3686.

\bibitem{Pan2010}
S.~J. Pan and Q.~Yang, ``A survey on transfer learning,'' \emph{{IEEE} Trans.
  on Knowledge and Data Engineering}, vol.~22, no.~10, pp. 1345--1359, 2010.

\bibitem{Pedrycz1998}
W.~Pedrycz, ``Fuzzy set technology in knowledge discovery,'' \emph{Fuzzy Sets
  and Systems}, vol.~98, pp. 279--290, 1998.

\bibitem{Quanz2009}
B.~Quanz and J.~Huan, ``Large margin transductive transfer learning,'' in
  \emph{Proc. 18th {ACM} Conf. on Information and Knowledge Management
  ({CIKM})}, Hong Kong, November 2009.

\bibitem{Ragin2000}
C.~C. Ragin, \emph{Fuzzy-set social science}.\hskip 1em plus 0.5em minus
  0.4em\relax Chicago, IL: The University of Chicago Press, 2000.

\bibitem{Sagberg2004}
F.~Sagberg, P.~Jackson, H.-P. Kruger, A.~Muzer, and A.~Williams, ``Fatigue,
  sleepiness and reduced alertness as risk factors in driving,'' Institute of
  Transport Economics, Oslo, Tech. Rep. TOI Report 739/2004, 2004.

\bibitem{Samek2013}
W.~Samek, F.~Meinecke, and K.-R. Muller, ``Transferring subspaces between
  subjects in brain-computer interfacing,'' \emph{{IEEE} Trans. on Biomedical
  Engineering}, vol.~60, no.~8, pp. 2289--2298, 2013.

\bibitem{Settles2009}
B.~Settles, ``Active learning literature survey,'' University of
  Wisconsin--Madison, Computer Sciences Technical Report 1648, 2009.

\bibitem{Spuler2012}
M.~Spuler, W.~Rosenstiel, and M.~Bogdan, ``Principal component based covariate
  shift adaption to reduce non-stationarity in a {MEG}-based brain-computer
  interface,'' \emph{{EURASIP} Journal on Advances in Signal Processing}, vol.
  2012, no.~1, pp. 1--7, 2012.

\bibitem{Tu2012b}
W.~Tu and S.~Sun, ``Dynamical ensemble learning with model-friendly classifiers
  for domain adaptation,'' in \emph{Proc. 21st Int'l Conf. on Pattern
  Recognition ({ICPR})}, Tsukuba, Japan, November 2012.

\bibitem{Tu2012}
W.~Tu and S.~Sun, ``A subject transfer framework for {EEG} classification,''
  \emph{Neurocomputing}, vol.~82, pp. 109--116, 2012.

\bibitem{NHTSA2011}
Traffic safety facts crash stats: drowsy driving. US Department of
  Transportation, National Highway Traffic Safety Administration. Washington,
  DC. [Online]. Available: \url{http://www-nrd.nhtsa.dot.gov/pubs/811449.pdf}

\bibitem{Erp2012}
J.~van Erp, F.~Lotte, and M.~Tangermann, ``Brain-computer interfaces: Beyond
  medical applications,'' \emph{Computer}, vol.~45, no.~4, pp. 26--34, 2012.

\bibitem{Versaci2003}
M.~Versaci and F.~C. Morabito, ``Fuzzy time series approach for disruption
  prediction in {Tokamak} reactors,'' \emph{{IEEE} Trans. on Magnetics},
  vol.~39, no.~3, pp. 1503--1506, 2003.

\bibitem{Vidaurre2011}
C.~Vidaurre, M.~Kawanabe, P.~V. Bunau, B.~Blankertz, and K.~Muller, ``Toward
  unsupervised adaptation of {LDA} for brain-computer interfaces,''
  \emph{{IEEE} Trans. on Biomedical Engineering}, vol.~58, no.~3, pp. 587--597,
  2011.

\bibitem{Walpole2007}
R.~W. Walpole, R.~H. Myers, A.~Myers, and K.~Ye, \emph{Probability \&
  Statistics for Engineers and Scientists}, 8th~ed.\hskip 1em plus 0.5em minus
  0.4em\relax Upper Saddle River, NJ: Prentice-Hall, 2007.

\bibitem{Wang1997}
L.-X. Wang, \emph{A Course in Fuzzy Systems and Control}.\hskip 1em plus 0.5em
  minus 0.4em\relax Upper Saddle River, NJ: Prentice Hall, 1997.

\bibitem{Wang2015}
P.~Wang, J.~Lu, B.~Zhang, and Z.~Tang, ``A review on transfer learning for
  brain-computer interface classification,'' in \emph{Prof. 5th Int'l Conf. on
  Information Science and Technology (IC1ST)}, Changsha, China, April 2015.

\bibitem{Wei2015}
C.-S. Wei, Y.-P. Lin, Y.-T. Wang, T.-P. Jung, N.~Bigdely-Shamlo, and C.-T. Lin,
  ``Selective transfer learning for {EEG}-based drowsiness detection,'' in
  \emph{Proc. {IEEE} Int'l Conf. on Systems, Man and Cybernetics}, Hong Kong,
  October 2015.

\bibitem{Welch1967}
P.~Welch, ``The use of fast {F}ourier transform for the estimation of power
  spectra: A method based on time averaging over short, modified
  periodograms,'' \emph{{IEEE} Trans. on Audio Electroacoustics}, vol.~15, pp.
  70--73, 1967.

\bibitem{Wolpaw2012}
J.~Wolpaw and E.~W. Wolpaw, Eds., \emph{Brain-Computer Interfaces: Principles
  and Practice}.\hskip 1em plus 0.5em minus 0.4em\relax Oxford, UK: Oxford
  University Press, 2012.

\bibitem{drwuaBCI2015}
D.~Wu, C.-H. Chuang, and C.-T. Lin, ``Online driver's drowsiness estimation
  using domain adaptation with model fusion,'' in \emph{Proc. Int'l Conf. on
  Affective Computing and Intelligent Interaction}, Xi'an, China, September
  2015.

\bibitem{drwuSMC2014}
D.~Wu, B.~J. Lance, and V.~J. Lawhern, ``Active transfer learning for reducing
  calibration data in single-trial classification of visually-evoked
  potentials,'' in \emph{Proc. {IEEE} Int'l Conf. on Systems, Man, and
  Cybernetics}, San Diego, CA, October 2014.

\bibitem{drwuPLOS2013}
D.~Wu, B.~J. Lance, and T.~D. Parsons, ``Collaborative filtering for
  brain-computer interaction using transfer learning and active class
  selection,'' \emph{{PLoS ONE}}, 2013.

\bibitem{drwuEBMAL2016}
D.~Wu, V.~J. Lawhern, S.~Gordon, B.~J. Lance, and C.-T. Lin, ``Offline
  {EEG}-based driver drowsiness estimation using enhanced batch-mode active
  learning ({EBMAL}) for regression,'' in \emph{Proc. {IEEE} Int'l Conf. on
  Systems, Man and Cybernetics}, Budapest, Hungary, October 2016.

\bibitem{drwuSMLR2016}
D.~Wu, V.~J. Lawhern, S.~Gordon, B.~J. Lance, and C.-T. Lin, ``Spectral
  meta-learner for regression {(SMLR)} model aggregation: Towards
  calibrationless brain-computer interface ({BCI}),'' in \emph{Proc. {IEEE}
  Int'l Conf. on Systems, Man and Cybernetics}, Budapest, Hungary, October
  2016.

\bibitem{drwuTNSRE2016}
D.~Wu, V.~J. Lawhern, W.~D. Hairston, and B.~J. Lance, ``Switching {EEG}
  headsets made easy: {Reducing} offline calibration effort using active
  weighted adaptation regularization,'' \emph{{IEEE} Trans. on Neural Systems
  and Rehabilitation Engineering}, 2016, in press.

\bibitem{drwuACII2015}
D.~Wu, V.~J. Lawhern, and B.~J. Lance, ``Reducing {BCI} calibration effort in
  {RSVP} tasks using online weighted adaptation regularization with source
  domain selection,'' in \emph{Proc. Int'l Conf. on Affective Computing and
  Intelligent Interaction}, Xi'an, China, September 2015.

\bibitem{drwuSMC2015}
D.~Wu, V.~J. Lawhern, and B.~J. Lance, ``Reducing offline {BCI} calibration
  effort using weighted adaptation regularization with source domain
  selection,'' in \emph{Proc. {IEEE} Int'l Conf. on Systems, Man and
  Cybernetics}, Hong Kong, October 2015.

\bibitem{drwuLS2011}
D.~Wu and J.~M. Mendel, ``Linguistic summarization using {IF-THEN} rules and
  interval type-2 fuzzy sets,'' \emph{{IEEE} Trans. on Fuzzy Systems}, vol.~19,
  no.~1, pp. 136--151, 2011.

\bibitem{drwuTL2011}
D.~Wu and T.~D. Parsons, ``Inductive transfer learning for handling individual
  differences in affective computing,'' in \emph{Proc. 4th Int'l Conf. on
  Affective Computing and Intelligent Interaction}, vol.~2, Memphis, TN,
  October 2011, pp. 142--151.

\bibitem{drwuICME2010}
D.~Wu, T.~D. Parsons, E.~Mower, and S.~S. Narayanan, ``Speech emotion
  estimation in {3D} space,'' in \emph{Proc. {IEEE} Int'l Conf. on Multimedia
  \& Expo ({ICME})}, Singapore, July 2010, pp. 737--742.

\bibitem{drwuEAAI2006}
D.~Wu and W.~W. Tan, ``Genetic learning and performance evaluation of type-2
  fuzzy logic controllers,'' \emph{Engineering Applications of Artificial
  Intelligence}, vol.~19, no.~8, pp. 829--841, 2006.

\bibitem{Yager1996}
R.~Yager, ``Database discovery using fuzzy sets,'' \emph{International Journal
  of Intelligent Systems}, vol.~11, pp. 691--712, 1996.

\bibitem{Zadeh1965}
L.~A. Zadeh, ``Fuzzy sets,'' \emph{Information and Control}, vol.~8, pp.
  338--353, 1965.

\bibitem{Zhu2004}
X.~Zhu and X.~Wu, ``Class noise vs. attribute noise: A quantitative study of
  their impacts,'' \emph{Artificial Intelligence Review}, vol.~22, pp.
  177--210, 2004.

\end{thebibliography}
\end{document}